\documentclass{article} %

\PassOptionsToPackage{sort}{natbib} %
\usepackage{preprint,times}

\usepackage{graphicx}
\usepackage{wrapfig}
\usepackage{float}
\usepackage{flafter}
\usepackage{placeins}
\usepackage{amsmath,amssymb}
\usepackage{xspace}
\usepackage{booktabs}
\usepackage{xcolor}
\usepackage{colortbl}
\usepackage{tabularx}
\definecolor{tableband}{HTML}{F1F4F7}
\definecolor{methodwash}{HTML}{EAF5F3}
\definecolor{referencewash}{HTML}{F0F3F8}
\definecolor{tableink}{HTML}{405363}
\newcommand{\sdfont}{\fontsize{6}{7}\selectfont}
\makeatletter
\newcommand{\asfigure}{\def\@captype{figure}}
\makeatother
\usepackage{pifont}
\usepackage{tcolorbox}
\usepackage{titletoc}%
\definecolor{tacdteal}{HTML}{16877D}
\definecolor{roseink}{HTML}{AD5058}
\definecolor{amplifiedwash}{HTML}{FCF3F3}
\newtcolorbox{finding}{colback=methodwash,colframe=tableink,boxrule=0.5pt,arc=2.5pt,
  left=5pt,right=5pt,top=3pt,bottom=3pt,before skip=7pt,after skip=7pt,
  before upper={\hyphenpenalty=10000 \exhyphenpenalty=10000}}%
\newcommand{\circnum}[1]{\textcolor{tacdteal}{\ding{\numexpr201+#1\relax}}}
\titlecontents{ssection}[1.5em]{\addvspace{4pt plus 5pt}\bfseries}
  {\contentslabel{1.5em}}{\hspace*{-1.5em}}{\hfill\contentspage}
\titlecontents{ssubsection}[3.7em]{}
  {\contentslabel{2.2em}}{\hspace*{-2.2em}}{\titlerule*[0.6em]{.}\contentspage}
\DeclareRobustCommand{\suppgroupline}[1]{\par\addvspace{12pt plus 12pt}\noindent
  {\footnotesize\bfseries\color{tacdteal}\MakeUppercase{#1}}\hspace{0.6em}%
  {\color{tacdteal!40}\leaders\hrule height 3.1pt depth -2.7pt\hfill}\par\nobreak\vspace{1pt}}
\newcommand{\suppgroup}[1]{\addtocontents{toc}{\protect\suppgroupline{#1}}}
\newcommand{\rng}[1]{{\footnotesize\textcolor{tableink}{[#1]}}}
\newcommand{\sd}[1]{\makebox[18pt][l]{\,{\sdfont$\pm$#1}}}
\newcommand{\apptab}{\scriptsize\setlength{\tabcolsep}{3pt}\renewcommand{\arraystretch}{1.0}}
\usepackage{pgfplots}
\pgfplotsset{compat=1.16}
\usepackage{multirow}
\usepackage{enumitem}
\usepackage{url}
\newcommand{\tacd}{\texorpdfstring{TACD}{TACD}\xspace}

\newcommand{\teacher}{v_{\mathrm{tea}}}
\newcommand{\stud}{v_{\theta}}
\newcommand{\xhat}{\hat{x}_1}

\definecolor{linkblue}{rgb}{0.12,0.46,0.70}
\usepackage[breaklinks,colorlinks,allcolors=linkblue]{hyperref}
\usepackage[capitalize]{cleveref}
\usepackage[ruled,vlined]{algorithm2e}
\usepackage{amsthm}

\crefname{remark}{Remark}{Remarks}
\Crefname{remark}{Remark}{Remarks}
\crefname{algocf}{Algorithm}{Algorithms}
\Crefname{algocf}{Algorithm}{Algorithms}

\title{TACD: Distilling Efficient Text-to-Motion\\
Models via Terminal Amplification Control}
\newcommand{\tacdline}[1]{\noindent\makebox[\textwidth]{\hspace*{\fill}#1\hspace*{\fill}}\par}
\author{\hspace*{-\tabcolsep}\makebox[0pt][l]{\parbox[t]{\textwidth}{%
\tacdline{Wei-Jin Huang$^{1}$\hfill Yuan-Ming Li$^{1}$\hfill Kun-Yu Lin$^{2}$\hfill
  Wang Luo$^{1}$\hfill Yinlin Zhu$^{1}$\hfill Yue Yu$^{3}$}
\vspace{1pt}
\tacdline{Shenghao Ye$^{4}$\hfill Junbin Yuan$^{1}$\hfill Fa-Ting Hong$^{5}$\hfill
  Qing Zhang$^{1}$\hfill Wei-Shi Zheng$^{1,\dagger}$}
\vspace{4pt}
{\normalfont\fontsize{8.5}{10}\selectfont
\tacdline{$^{1}$Sun Yat-sen University\hfill $^{2}$Nanyang Technological University\hfill
  $^{3}$Wuhan University}
\tacdline{$^{4}$University of Science and Technology of China\hfill
  $^{5}$The Hong Kong University of Science and Technology}
\vspace{2pt}
\tacdline{\ttfamily huangwj235@mail2.sysu.edu.cn\hfill wszheng@ieee.org}
\vspace{1pt}
\tacdline{Project page: \href{https://vkgo.github.io/TACD/}{\ttfamily https://vkgo.github.io/TACD/}}}}}}

\begin{document}
\raggedbottom
\maketitle
{\renewcommand{\thefootnote}{\fnsymbol{footnote}}\footnotetext[2]{Corresponding author.}}
\vspace{6pt}%

\lhead{Preprint}

\begin{figure}[H]
  \centering
  \vspace{-10mm}
  \includegraphics[width=\linewidth]{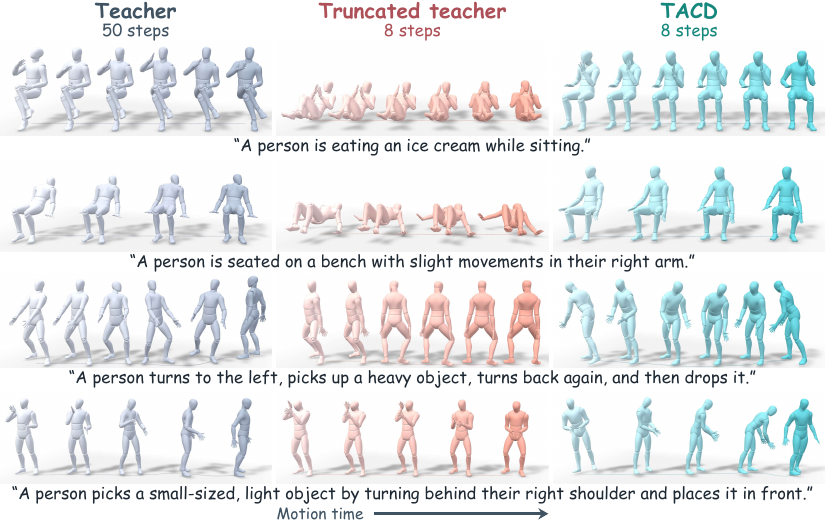}
  \vspace{-5.4mm}
  \caption{\textbf{Motion examples.} Teacher, the same teacher truncated to fewer steps, and TACD. Rows
  share prompt, noise, frames, and camera (\cref{app:teaser3}).}
  \label{fig:teaser3}
\end{figure}

\vspace{-3mm}%
\begin{abstract}
Recent text-to-motion models have improved motion quality and instruction following, yet many-step denoising and large model components make deployment slow and memory-intensive.
We present Terminal-Amplification-Controlled Distillation (TACD), an on-policy approach for training efficient motion generators from text prompts and pretrained teachers, without real-motion training data.
Building on segmented on-policy flow distillation, we supervise clean-motion predictions along student-generated trajectories.
We identify a failure mode in which velocity matching on a fixed supervision grid repeatedly overweights errors near the denoising endpoint, degrading few-step generation.
TACD ties the latest teacher query to the student's step size, bounding the effective loss weights in clean-motion space without changing inference.
Experiments on HumanML3D and KIT-ML demonstrate improved few-step generation, including a 58\% reduction in eight-step HY-Motion student FID relative to distillation without this bound.
For diffusion teachers, the endpoint-matching form of TACD yields four-step students with lower FID and matched or improved text--motion retrieval relative to their 50-step teachers on HumanML3D.
On HY-Motion and Kimodo, eight-step students with compact components achieve $7.7$--$11.9\times$ end-to-end speedups and reduce peak GPU memory by $3.8$--$6.7\times$ relative to their teachers.

\end{abstract}

\section{Introduction}
\label{sec:intro}

Text-to-motion generation provides an intuitive interface for creating human motion from
natural-language descriptions, with applications in character animation and interactive virtual
environments. Recent diffusion and flow models have made substantial progress in motion synthesis
and instruction following~\citep{tevet2023mdm,hymotion2025}. Yet generation quality alone is
insufficient for interactive use: models must also respond promptly to user instructions and operate
within limited device memory. These requirements are difficult to meet when generation involves
repeated evaluations of a large motion model supported by a heavy text encoder.

Existing acceleration methods replace long denoising procedures with few-step generation.
MotionLCM~\citep{dai2024motionlcm} applies latent consistency distillation to motion generation,
while MotionPCM~\citep{motionpcm2025} introduces phased consistency to improve synthesis quality and
efficiency. However,
motion-distillation pipelines such as MotionLCM use real motion examples during training, which may
be unavailable when adapting a released generator trained on an inaccessible corpus. Moreover,
reducing sampling steps alone does not reduce the size of the motion backbone or its conditioning
encoder. This motivates learning few-step, compact students from available text prompts and
pretrained teachers, without requiring real-motion training
data~\citep{gu2023boot,zhou2024sidlsg,xiang2025dkdm}.

Teacher-guided on-policy distillation offers a natural starting point by obtaining supervision
along trajectories generated by the student. In particular, $\pi$-Flow~\citep{piflow2025} learns a
policy that can be queried between student network evaluations and matches it to the teacher along
its own trajectory. This construction provides teacher feedback on states encountered during
generation. Nevertheless, selecting these states does not determine how strongly their prediction
errors should be weighted. In our motion-distillation experiments, a student trained with this form
of teacher supervision can still exhibit substantial few-step degradation. The central question of
this paper is therefore how strongly each supervised state should count.

\begin{wrapfigure}{r}{0.44\linewidth}
  \vspace{-10mm}
  \centering
  \includegraphics[width=\linewidth]{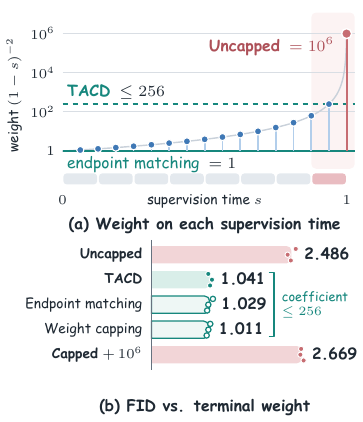}
  \vspace{-9pt}
  \vspace{-5mm}
  \caption{\textbf{Controlling terminal amplification improves few-step motion distillation.}
  \textbf{(a)}~The known weight \mbox{$(1-s)^{-2}$} on our supervision grid. \textbf{(b)}~Bounding it
  improves FID; hollow bars enforce the same bound on the original grid (dots: seeds).}
  \label{fig:hero}
  \vspace{-5mm}
\end{wrapfigure}

Our analysis reveals a problematic interaction between the prediction space and the supervision
schedule. When clean-motion predictions are trained through velocity matching, their errors receive
increasingly large weights near the denoising endpoint. Although this conversion-induced weighting
has been studied under random timestep sampling~\citep{hong2026alignment}, a fixed supervision grid
queries the same near-terminal time at every training update. The resulting
amplification is therefore a persistent component of the loss rather than an occasional heavily
weighted sample. Controlled interventions show that bounding this weight restores few-step motion
quality and that restoring it reverses the gain (\cref{fig:hero}).

We introduce \emph{Terminal-Amplification-Controlled Distillation} (TACD), an on-policy distillation
method for efficient text-to-motion generation. Building on $\pi$-Flow's single-prediction-per-segment
construction, the student predicts the entire clean motion at the start of each denoising segment
and learns from teacher targets along student-generated trajectories. TACD regulates the resulting
supervision weights to prevent excessive terminal amplification: it ties the latest supervision
time to the student's step size, which bounds every weight in closed form. Controlled comparisons
show that other ways of enforcing the same bound recover comparable quality. The same construction
extends to diffusion teachers, for which TACD takes its endpoint-matching form.

On HumanML3D, terminal amplification control reduces the eight-step student's FID from $2.486$ to
$1.041$, and restoring the terminal coefficient alone raises it again to $2.669$. Beyond this flow
intervention, four-step endpoint-matching students of two diffusion teachers match or improve their
teachers on FID and text--motion retrieval on this benchmark, and on KIT-ML and Kimodo~\citep{rempe2026kimodo} the students
improve on the teacher truncated to the same step count. Under the SSAE vision--language judge~\citep{hymotion2025} on 640 free-form
prompts, the eight-step HY-Motion student matches its 50-step teacher and leads
seven of ten open-weight HumanML3D models (\cref{tab:ssae}); in blind human rankings, its win rate against the teacher
exceeds those of the four-step MotionLCM-V2 and MLCT by $10$ and $17$ percentage points (\cref{tab:human}). Efficiency studies on two large generators
(\cref{fig:cost}) show eight-step students that are \textbf{$\mathbf{7.7\times}$ (HY-Motion) and
$\mathbf{11.9\times}$ (Kimodo) faster} end to end than their teachers, with smaller motion backbones and text
encoders ($7.0\times$ and $4.7\times$ fewer parameters) reducing \textbf{GPU memory by $\mathbf{6.7\times}$ and
$\mathbf{3.8\times}$}.

Our contributions are threefold: (1) we identify terminal amplification in segmented
velocity-matching distillation, where a fixed supervision grid places a persistent, orders-of-magnitude
larger weight at the same near-terminal supervision time, and show through controlled interventions that it degrades few-step
motion quality; (2) we introduce TACD, which ties the latest teacher query to the student's step size
and thereby bounds every supervision weight, and extend it to diffusion teachers through its
endpoint-matching form; (3) on HumanML3D, KIT-ML, and Kimodo, few-step TACD students improve on the
teacher truncated to their step count, and, combined with compact backbones and text encoders,
eight-step students run $7.7$--$11.9\times$ faster end to end and use $3.8$--$6.7\times$ less peak GPU memory than their
teachers.

\section{Related work}
\label{sec:related}

\paragraph{Efficient and data-free motion distillation.}
Text-to-motion generators include VAEs~\citep{petrovich2022temos}, token models
\citep{guo2022t2mgpt,guo2024momask}, diffusion models~\citep{tevet2023mdm,chen2023mld},
and flow matching~\citep{lipman2023flowmatching}, as in HY-Motion~\citep{hymotion2025}.
Sampling acceleration includes progressive and consistency distillation
\citep{salimans2022progressive,song2023consistency,luo2023lcm}, distribution and moment
matching~\citep{yin2024dmd,yin2024dmd2,salimans2024moment,luo2025tdm}, and average-velocity
modeling~\citep{geng2025meanflow}. MotionLCM~\citep{dai2024motionlcm} uses real motions
during distillation; MotionPCM~\citep{motionpcm2025} combines phased consistency and
adversarial training. We adopt teacher-output supervision without real-motion training targets
\citep{gu2023boot,zhou2024sidlsg,xiang2025dkdm}, and evaluate both sampling and model costs; solver and
compression methods address complementary inference costs
\citep{lu2022dpmsolver,li2023qdiffusion,he2024efficientdm,sun2024tmpqdm,amin2025diffpro,du2026wanquant}.

\paragraph{Policy-based generative distillation.}
$\pi$-Flow~\citep{piflow2025} constructs a queryable DX velocity policy from predicted
clean-sample estimates. We use its single-knot specialization: one estimate defines a
segment, and teacher queries supervise the student-generated trajectory. Other on-policy
distillers study arbitrary-interval flow maps~\citep{anyflow2026}, flow-matching distillation
\citep{flowopd2026}, or black-box adversarial transfer~\citep{afd2026}. Our focus is how the
endpoint parameterization interacts with the finite supervision grid of a few-step student.

\paragraph{Prediction--loss alignment and temporal supervision.}
EDM~\citep{karras2022edm} and Min-SNR~\citep{minsnr2023} study prediction parameterizations
and their induced loss weights. \citet{hong2026alignment} identify the inverse-square
endpoint-to-velocity factor, analyze terminal amplification under random timestep sampling,
and study prediction-space losses and capped weights. JiT~\citep{li2025jit} clips the
conversion denominator; FlowCTS~\citep{flowcts2026} identifies concentration from
transition-kernel KL weighting. Random time sampling draws a heavily weighted time only occasionally;
on a fixed grid over detached student rollouts the same near-terminal supervision time recurs at every update,
and its weight becomes a fixed loss coefficient. This permits separate interventions on supervision time
and coefficient and a bound tied to the student's step size; TACD's endpoint-matching form for
diffusion teachers applies prediction--loss alignment.

\FloatBarrier
\section{Method}
\label{sec:method}
TACD is a supervision design for segmented on-policy distillation. In each denoising segment the
student predicts the complete clean motion once; this prediction fixes the student's states within
the segment, and the frozen teacher, queried at those states during training only, supplies the targets that correct it. We first formulate this construction for flow teachers and analyze the effective
supervision weights induced by velocity matching. We then introduce terminal amplification control
and extend its endpoint-matching formulation to diffusion teachers.

\begin{figure}[!htbp]
  \centering
  \includegraphics[width=\linewidth]{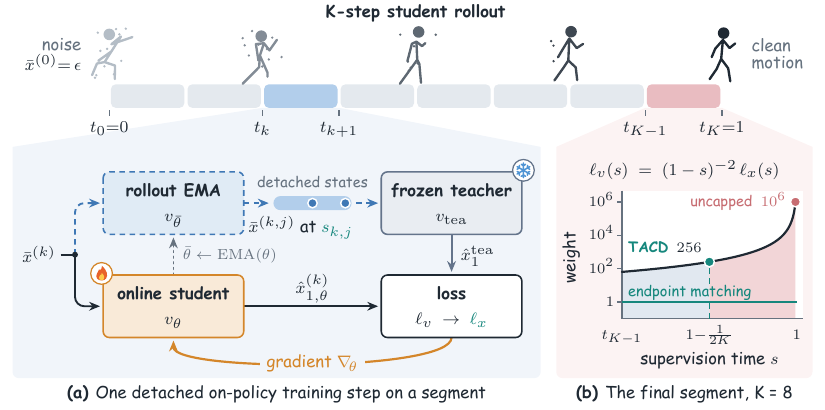}
  \vspace{-7mm}
  \caption{\textbf{Flow training and terminal amplification control.}
  The rollout EMA generates detached supervision states, while gradients update the online student.
  TACD moves the supervision times $s_{k,j}$ so that every coefficient is bounded; endpoint matching
  ($\ell_x$) enforces the same bound at the original queries (teal marks what each changes). Both
  preserve the inference schedule.}
  \vspace{-2mm}
  \label{fig:controls}
\end{figure}

\subsection{Segmented on-policy distillation}
\label{sec:setting}
\label{sec:onpolicy}
Given text $c$ and noise, the student generates a complete motion in $K$ evaluations, without
real-motion training targets. Flow time runs from noise at $t=0$ to clean motion at $t=1$, and $x_t$
is the entire motion at flow time $t$, not a physical frame. The \emph{sampling grid}
$0=t_0<\cdots<t_K=1$ fixes the $K$ student evaluations and splits denoising into the segments
$[t_k,t_{k+1}]$. Following the single-knot DX parameterization of $\pi$-Flow~\citep{piflow2025}, one
clean-motion prediction defines the student policy over each segment: at the segment start
$x^{(k)}$, the student's velocity output $\stud$ gives
\begin{align}
\hat x_{1,\theta}^{(k)}&=x^{(k)}+(1-t_k)\stud(x^{(k)},t_k;c),
\label{eq:fast_endpoint}\\
\pi_\theta^{(k)}(x,t)&=\frac{\hat x_{1,\theta}^{(k)}-x}{1-t},\qquad t\in[t_k,t_{k+1}].
\label{eq:segment_policy}
\end{align}
The prediction estimates the clean motion at $t=1$, not the state at the next boundary; holding it
fixed within the segment yields the update
\begin{equation}
x^{(k+1)}=x^{(k)}+\frac{t_{k+1}-t_k}{1-t_k}
\left(\hat x_{1,\theta}^{(k)}-x^{(k)}\right).
\label{eq:inference_update}
\end{equation}

\paragraph{Supervision along the student rollout.}
\label{sec:detached_training}
Teacher supervision is obtained on states that the student itself visits. An EMA copy
$\bar\theta$ of the student generates the rollout: starting from noise $\bar x^{(0)}$, it applies
\cref{eq:fast_endpoint,eq:inference_update} to produce the segment starts $\bar x^{(k)}$. Within
each segment we place $M$ supervision times
\begin{equation}
s_{k,j}(\delta)=t_k+\frac{j}{M}
\left[\min(t_{k+1},1-\delta)-t_k\right],\qquad j=1,\ldots,M,
\label{eq:cap}
\end{equation}
where the terminal margin $\delta>0$ keeps the last query below $t=1$, at which
\cref{eq:segment_policy} is undefined. The \emph{supervision grid} $\mathcal S=\{s_{k,j}\}$ is used
only during training and is distinct from the sampling grid, which always ends at $t_K=1$. The
rollout states at the supervision times are
\begin{equation}
\bar x^{(k,j)}=\operatorname{sg}\!\left[
\bar x^{(k)}+\frac{s_{k,j}-t_k}{1-t_k}
\left(\hat x_{1,\bar\theta}^{(k)}-\bar x^{(k)}\right)\right],
\label{eq:ema_state}
\end{equation}
with $\operatorname{sg}[\cdot]$ the stop-gradient. For the loss, the online student computes its prediction
$\hat x_{1,\theta}^{(k)}$ once at each detached segment start and reuses it for the segment's $M$
queries; the frozen teacher velocity $\teacher$ is queried at the supervision states. Velocity
matching penalizes, at each query, the residual
\begin{equation}
\ell_v(s_{k,j})=\left\|\pi_\theta^{(k)}(\bar x^{(k,j)},s_{k,j})
-\teacher(\bar x^{(k,j)},s_{k,j};c)\right\|_m^2 ,
\qquad
\widetilde{\mathcal L}_{v}=\sum_{k=0}^{K-1}\frac{w_k}{M}\sum_{j=1}^{M}\ell_v(s_{k,j}),
\label{eq:lagged}
\end{equation}
where $w_k$ are segment weights and $\|\cdot\|_m^2$ is the squared error averaged over valid frames
and channels. Rollout states and teacher targets are constants for the update, so gradients pass
only through the online predictions at the segment starts. This construction fixes \emph{where}
supervision is obtained; the matching space sets how strongly each query counts.

\subsection{Terminal amplification in velocity matching}
\label{sec:singularity}
The teacher's instantaneous velocity implies a clean-motion estimate,
\begin{equation}
\hat x_1^{\mathrm{tea}}(x,t;c)=x+(1-t)\teacher(x,t;c).
\label{eq:teacher_endpoint}
\end{equation}
Substituting it into the residual of \cref{eq:lagged} at a supervision state $\bar x$ and time $s$
gives
\begin{equation}
\pi_\theta^{(k)}(\bar x,s)-\teacher(\bar x,s;c)
=\frac{\hat x_{1,\theta}^{(k)}-\hat x_1^{\mathrm{tea}}(\bar x,s;c)}{1-s},
\qquad\text{hence}\qquad
\ell_v(s)=(1-s)^{-2}\,\ell_x(s),
\label{eq:conversion}
\end{equation}
with $\ell_x(s)=\|\hat x_{1,\theta}^{(k)}-\hat x_1^{\mathrm{tea}}(\bar x,s;c)\|_m^2$ the clean-motion
loss: velocity matching is clean-motion matching with an inverse-square time-dependent
weight~\citep{hong2026alignment}. The objective of \cref{eq:lagged} is therefore
\begin{equation}
\mathcal L_{\mathcal S,b}
=\sum_{k=0}^{K-1}\frac{w_k}{M}\sum_{j=1}^{M}b_{k,j}
\left\|\hat x_{1,\theta}^{(k)}
-\operatorname{sg}\!\left[\hat x_1^{\mathrm{tea}}(\bar x^{(k,j)},s_{k,j};c)\right]\right\|_m^2,
\qquad b_{k,j}=(1-s_{k,j})^{-2},
\label{eq:unified}
\end{equation}
where the segment weights $w_k$ are a design choice but the coefficients $b_{k,j}$ are not: they are
induced by matching in velocity space.

The last query of the supervision grid sits at $s_{K-1,M}=1-\delta$ and receives the
\emph{terminal coefficient} $b_{K-1,M}=\delta^{-2}$. On a uniform sampling grid every other query
satisfies $1-s_{k,j}\ge1/(MK)$, so its coefficient is at most $(MK)^2$: a small numerical margin
makes one coefficient exceed all others by orders of magnitude ($\delta=10^{-3}$ gives $10^6$).
Because the supervision grid is reused at every update, this weight returns at the same supervision time in every
update, whatever state the rollout reaches there, instead of being met occasionally as under random time sampling. We call this persistent
overweighting \emph{terminal amplification}. A large coefficient does not by itself make its term
dominate the update, since each query's gradient share also depends on its residual and on the
student Jacobian (\cref{eq:gradient_conversion}); \cref{sec:onestate} measures gradient shares and
intervenes on the coefficient directly.

\subsection{Terminal amplification control}
\label{sec:tacd_repairs}
TACD bounds every coefficient $b_{k,j}$ in \cref{eq:unified} by a value tied to the student's step
size. It retains the sampling grid, the inference update, and the training budget, changes only the
supervision used during training, and adds no network module or training stage. For a flow teacher
it ties the terminal margin in \cref{eq:cap} to the step $\Delta t=1/K$ of the uniform sampling grid
$t_k=k/K$:
\begin{equation}
\delta=\frac{\Delta t}{2}=\frac{1}{2K},
\qquad\text{so that}\qquad
\max_{k,j}b_{k,j}=\delta^{-2}=\frac{4}{\Delta t^{2}}=4K^2 .
\label{eq:midpoint}
\end{equation}
The latest teacher query, the upper end of the final supervision interval, then sits at the midpoint
of the final student step, and \cref{eq:cap} spaces the final
segment's $M$ queries uniformly within this shortened interval, so the teacher is never queried next
to the singular time, while the sampling grid still ends at $t_K=1$. The terminal coefficient has the
same order as the others on the grid, and the rule is closed-form in the step count, with no tuned
threshold. Grid capping is
TACD's rule for every flow result: it bounds the weights while keeping the velocity loss of
\cref{eq:unified} unchanged.

Two other enforcements meet the same bound at the original queries, each with a different loss; they
serve as controls in \cref{sec:repair}, and one of them, endpoint matching, is the form TACD takes for
diffusion teachers (\cref{sec:ddim_instance}). Weight capping truncates the coefficient,
$b_{k,j}=\min\{(1-s_{k,j})^{-2},\,4K^2\}$, the truncated-weight form of \citet{hong2026alignment}
related to denominator clipping~\citep{li2025jit}, which with $M=2$ changes a single term of
\cref{eq:unified}; endpoint matching sets $b_{k,j}\equiv1$, which matches the teacher's clean-motion estimates
directly and applies prediction--loss alignment~\citep{hong2026alignment} to fixed-grid on-policy
supervision. The three enforcements differ in how they meet the bound: grid capping moves the final-segment queries,
weight capping lowers one coefficient, and endpoint matching removes the conversion factor at every
query.

\subsection{Extension to diffusion teachers}
\label{sec:ddim_instance}
For a diffusion teacher the student again predicts clean motion at each segment start and learns
from frozen-teacher targets on detached rollout states, with two changes: the states follow
deterministic DDIM sub-trajectories, and the loss is \cref{eq:unified} with $b_{k,j}\equiv1$. Let $\tau_0>\cdots>\tau_K$ be the teacher sampler's $K$-step timestep schedule
with cumulative signal level $\bar\alpha_\tau$, and $\hat x_{0,\theta}^{(k)}$ the student's
clean-motion prediction at the segment start $\bar x^{(k)}$ ($x_0$ is the flow case's $x_1$
under reversed time). Holding the rollout prediction
$\hat x_{0,\bar\theta}^{(k)}$ fixed over the segment also fixes the noise $\bar\epsilon^{(k)}$ it
implies at $\bar x^{(k)}$, and the DDIM states at $M$ supervision times $\tau_{k,j}$ within the
segment,
\begin{equation}
\bar x^{(k,j)}=\operatorname{sg}\!\left[\sqrt{\bar\alpha_{\tau_{k,j}}}\,\hat x_{0,\bar\theta}^{(k)}
+\sqrt{1-\bar\alpha_{\tau_{k,j}}}\;\bar\epsilon^{(k)}\right],
\label{eq:ddim_state}
\end{equation}
take the place of \cref{eq:ema_state}. The target is the frozen teacher's clean-motion estimate
$\hat x_0^{\mathrm{tea}}$ (its clean-sample output, or the DDIM predicted sample of a
noise-prediction teacher), computed with the teacher sampler's guidance (\cref{eq:onx0} in \cref{sec:exactobj}). Written in
clean-motion space from the start, this loss carries no $(1-s)^{-2}$ factor; matching noise predictions on
the same states would instead weight the clean-motion error by the signal-to-noise
ratio~\citep{salimans2022progressive,minsnr2023}, which also grows without bound toward the clean end
(\cref{eq:snr}). Endpoint matching thus carries no time-dependent weight.
Grid capping acts on the flow conversion factor and is specific to flow teachers; for diffusion
teachers TACD denotes this endpoint-matching form. Teacher-specific segment weights and rollout
settings are reported with each experiment; pseudocode and implementation settings are in
\cref{sec:exactobj,app:training_settings}.

\FloatBarrier %
\raggedbottom %
\begin{table}[t]
\scriptsize
\renewcommand{\arraystretch}{1.025}
\setlength{\tabcolsep}{3pt}
\begin{minipage}[t]{0.64\linewidth}
\vspace{0pt}
\centering
\resizebox{\linewidth}{!}{%
\begin{tabular}{@{}c@{\hspace{3pt}}lccccccc@{}}
\toprule
& Method & NFE & R@1$\uparrow$ & R@3$\uparrow$ & MM-Dist$\downarrow$ & FID$\downarrow$ & Div$\rightarrow$ & MMod$\uparrow$ \\
\midrule
& \multicolumn{8}{@{}l}{\emph{Published generators (real-motion training) and real-motion reference}} \\
& Real motions~\citep{guo2022humanml3d} & -- & .511 & .797 & 2.974 & .002 & 9.503 & -- \\
& MDM (fixed eval.)~\citep{tevet2023mdm} & $\sim$1000 & .418 & .707 & 3.63 & .489 & 9.45 & 2.87 \\
& T2M-GPT~\citep{guo2022t2mgpt} & -- & .492 & .775 & 3.121 & .141 & 9.722 & 1.831 \\
& MoMask (full)~\citep{guo2024momask} & -- & .521 & .807 & 2.958 & .045 & -- & 1.241 \\
& IRG-MotionLLM S3~\citep{irg_motionllm} & -- & .535 & .820 & 2.785 & .242 & 9.900 & -- \\
& MotionHiFlow~\citep{motionhiflow} & -- & .563 & .843 & 2.691 & .032 & 9.504 & -- \\
& MoGeFlow-L~\citep{moge_flow} & -- & .592 & .873 & 2.599 & .058 & 9.981 & -- \\
\cmidrule(l){2-9}
& \multicolumn{8}{@{}l}{\emph{Published few-step generators (real-motion training; $^{\ast}$released checkpoint, our evaluation)}} \\
& MotionLCM-V2 student$^{\ast}$~\citep{dai2024motionlcmv2} & 4 & .554 & .841 & 2.760 & .072 & 9.54 & 1.72 \\
& MLCT~\citep{hu2024mlct} & 4 & .537 & .826 & 2.819 & .060 & 9.545 & 2.571 \\
\multirow{-12}{*}{\rotatebox[origin=c]{90}{\emph{Real-motion training}}}& MotionPCM~\citep{motionpcm2025} & 4 & .559 & .842 & 2.716 & .030 & 9.575 & 1.714 \\
\midrule
& \multicolumn{8}{@{}l}{\emph{Our evaluation: HY-Motion references}} \\
& HY-Motion teacher & 50 & .455 & .743 & 3.471 & .541 & 9.13 & 3.28 \\
& HY-Motion Lite (released) & 50 & .435 & .728 & 3.575 & .726 & 8.95 & -- \\
& Truncated teacher & 8 & .378 & .652 & 4.225 & 3.250 & 8.24 & -- \\
\cmidrule(l){2-9}
& \multicolumn{8}{@{}l}{\emph{Controlled Lite students, three-seed means: Qwen3-0.6B encoder, teacher-only supervision}} \\
& Uncapped student (DX, $N{=}1$) & 8 & .402 & .708 & 3.703 & 2.486 & 8.01 & \textbf{3.994} \\
\rowcolor{methodwash}
\cellcolor{white}& \textbf{\tacd} & 8 & \textbf{.479} & \textbf{.782} & \textbf{3.192} & \textbf{1.041} & 8.92 & 3.541 \\
\cmidrule(l){2-9}
& \multicolumn{8}{@{}l}{\emph{Non-adversarial ports to the Lite student, three-seed means: teacher-only supervision}} \\
& DMD port & 8 & .426 & .730 & 3.493 & 1.829 & 8.22 & 3.023 \\
& Consistency port & 8 & .394 & .699 & 3.754 & 2.908 & 7.77 & 3.803 \\
\multirow{-11}{*}{\rotatebox[origin=c]{90}{\emph{HY-Motion teacher}}}& Reflow port & 8 & .370 & .660 & 4.063 & 3.341 & 8.07 & 3.886 \\
\bottomrule
\end{tabular}}
\caption{\textbf{HumanML3D.} At eight steps with the HY-Motion teacher, \tacd\ is best in R@1, R@3, MM-Dist, and FID.}
\label{tab:matched_lite}
\label{tab:published_hml}
\end{minipage}\hfill
\begin{minipage}[t]{0.34\linewidth}
\vspace{0pt}
\centering
\includegraphics[width=\linewidth]{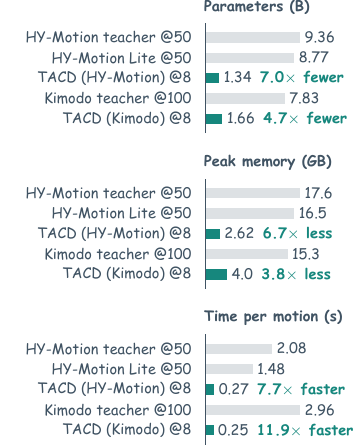}
\asfigure
\caption{\textbf{Deployment cost}, RTX L40, batch 1 (\cref{sec:latency}); $\times$: vs.\ the teacher; @$N$: $N$ steps.}
\label{fig:cost}
\end{minipage}
\end{table}

\section{Experiments}
\label{sec:exp}
On HumanML3D, the HY-Motion flow teacher tests whether \tacd\ improves few-step students
(\cref{sec:overall}) and whether the terminal coefficient is the cause (\cref{sec:onestate}).
\Cref{sec:extension} tests the endpoint-matching form with the MLD and MotionLCM-V2 diffusion teachers
and adds KIT-ML (MDM teacher) and Kimodo (compact encoder); \cref{sec:coupled} measures deployment cost.
\subsection{Experimental setup}
\label{sec:exp_setup}
\label{sec:compact}
HumanML3D and KIT-ML use dataset-specific Guo evaluators and real-motion references. The main flow student, \tacd\ (HY-Motion), pairs the 0.46B motion backbone of the released HY-Motion Lite with frozen
Qwen3-0.6B conditioning and a trainable projection. Controls share initialization, 6k prompts,
8k updates, eight-step inference, CFG 5, and final evaluation-EMA checkpoints; only supervision
changes. They use $M=2$ supervision times per segment, segment weights $w_k=1$, and, on the
original supervision grid, the terminal margin $\delta=10^{-3}$. The diffusion students are
teacher-initialized, low-rank adapted, and trained with DDIM endpoint matching
(\cref{sec:ddim_instance}) at two, four, and eight steps. Training uses training-split captions as
text only, with teacher supervision and no real-motion targets; evaluation uses the test split.
\Cref{tab:mardm} repeats the HY-Motion and MLD comparisons with the MARDM evaluator
\citep{meng2025mardm} (protocol in \cref{app:mardm}).

Kimodo~\citep{rempe2026kimodo} students keep its 0.28B backbone with LoRA and use its 8B encoder or a
trained-in 1B encoder, \tacd\ (Kimodo), evaluated on 607 captions with Kimodo's TMR evaluator. Full settings and evaluation details are in \cref{app:training_settings,app:kimodo}.

\paragraph{Metrics.} Following \citet{guo2022humanml3d}, FID compares generated and real motions in a
pretrained feature space, R-precision (R@$k$) and MM-Dist measure text--motion alignment, and Diversity (Div;
$\rightarrow$: closer to real is better) and MultiModality (MMod) measure variation across and within prompts;
Kimodo uses TMR-based retrieval and FID~\citep{petrovich2023tmr}. NFE counts guided denoising steps per motion
(sampling steps; a guided step counts once); definitions are in \cref{app:evaluation}.
\newsavebox{\ssaebox}%
\newlength{\ssaewidth}%
\begin{table}[t]
\scriptsize
\setlength{\tabcolsep}{3pt}
\renewcommand{\arraystretch}{1.0}
\renewcommand{\rng}[1]{\textcolor{tableink}{[#1]}}%
\setlength{\ssaewidth}{\dimexpr0.98\linewidth-0.595\linewidth\relax}%
\begin{minipage}[t]{0.595\linewidth}
\vspace{0pt}
\centering
\resizebox{\linewidth}{!}{%
\begin{tabular}{@{}lcccccccc@{}}
\toprule
Method & NFE & R@1$\uparrow$ & R@2$\uparrow$ & R@3$\uparrow$ & MM-Dist$\downarrow$ & FID$\downarrow$ & Div$\rightarrow$ & CLIP$\uparrow$ \\
\midrule
Real motions & -- & .505 & .702 & .801 & 3.241 & -- & 10.33 & .638 \\
\midrule
\multicolumn{9}{@{}l}{\emph{HY-Motion flow teacher; students at eight steps, three-seed means}} \\
HY-Motion teacher & 50 & .415 & .601 & .708 & 4.056 & 3.119 & 9.47 & .604 \\
HY-Motion Lite (released) & 50 & .394 & .578 & .685 & 4.243 & 3.741 & 9.29 & .589 \\
Truncated teacher & 8 & .369 & .544 & .646 & 4.721 & 6.816 & 8.92 & .510 \\
Uncapped student (DX, $N{=}1$) & 8 & .343 & .528 & .647 & 4.562 & 7.559 & 8.24 & .569 \\
\rowcolor{methodwash}
\textbf{\tacd} & 8 & .412 & .606 & .721 & 3.955 & 3.575 & 9.21 & .621 \\
DMD port & 8 & .371 & .559 & .678 & 4.299 & 5.789 & 8.67 & .591 \\
Consistency port & 8 & .325 & .505 & .624 & 4.697 & 7.317 & 8.11 & .563 \\
Reflow port & 8 & .321 & .492 & .604 & 4.980 & 10.473 & 8.21 & .526 \\
\midrule
\multicolumn{9}{@{}l}{\emph{MLD diffusion teacher; \tacd\ at four steps, three-seed mean}} \\
MLD teacher (DDIM) & 50 & .483 & .680 & .780 & 3.339 & .512 & 10.35 & .630 \\
Truncated MLD teacher & 4 & .447 & .645 & .754 & 3.556 & .735 & 10.31 & .599 \\
\rowcolor{methodwash}
\textbf{\tacd} & 4 & .507 & .703 & .801 & 3.215 & .228 & 10.70 & .639 \\
\bottomrule
\end{tabular}}
\caption{\textbf{MARDM evaluator} \citep{meng2025mardm}: \tacd\ leads the truncated teacher and every eight-step student in R@$k$, MM-Dist, FID, and CLIP score; four-step MLD \tacd\ matches real motions in R-precision.}
\label{tab:mardm}
\global\setbox\ssaebox=\hbox{%
\begin{minipage}[t]{\ssaewidth}
\vspace{0pt}
\centering
\renewcommand{\arraystretch}{0.970}%
\resizebox{\linewidth}{!}{%
\begin{tabular}{@{}lcc@{}}
\toprule
Model & SSAE$\uparrow$ & $\Delta$ vs.\ \tacd\ \rng{95\% CI} \\
\midrule
HY-Motion teacher @50 & .695 & $-.025$ \rng{$-.053$, $+.004$} \\
HY-Motion Lite (released) @50 & .692 & $-.028$ \rng{$-.057$, $+.001$} \\
Truncated teacher @8 & .587 & $-.133$ \rng{$-.166$, $-.100$} \\
Uncapped student @8 & .650 & $-.070$ \rng{$-.093$, $-.045$} \\
DMD port @8 & .672 & $-.048$ \rng{$-.071$, $-.024$} \\
Consistency port @8 & .700 & $-.020$ \rng{$-.044$, $+.006$} \\
Reflow port @8 & .590 & $-.130$ \rng{$-.156$, $-.102$} \\
\rowcolor{methodwash}
\textbf{\tacd} @8 & \textbf{.720} & -- \\
\midrule
\multicolumn{3}{@{}l}{\emph{Open-weight HumanML3D models, all rendered from joints}} \\
MDM @1000 & .588 & $-.112$ \rng{$-.141$, $-.080$} \\
MLD @50 & .641 & $-.058$ \rng{$-.089$, $-.026$} \\
T2M-GPT & .618 & $-.081$ \rng{$-.112$, $-.051$} \\
MoMask & .636 & $-.063$ \rng{$-.093$, $-.033$} \\
MARDM & .642 & $-.057$ \rng{$-.086$, $-.028$} \\
MotionHiFlow & .679 & $-.021$ \rng{$-.052$, $+.010$} \\
MoGeFlow-L & .716 & $+.017$ \rng{$-.013$, $+.046$} \\
MotionLCM @4 & .661 & $-.038$ \rng{$-.068$, $-.007$} \\
MLCT @4 & .666 & $-.033$ \rng{$-.062$, $-.004$} \\
MotionLCM-V2 @4 & .697 & $-.002$ \rng{$-.031$, $+.029$} \\
\rowcolor{methodwash}
\textbf{\tacd} @8, from joints & .699 & -- \\
\bottomrule
\end{tabular}}
\caption{\textbf{SSAE:} a VLM judge's yes-rate on questions about the prompted motion (\cref{app:ssae}); lower block: $\Delta$ vs.\ \tacd\ from joints.}
\label{tab:ssae}
\par\vspace{4pt}
\resizebox{\linewidth}{!}{%
\begin{tabular}{@{}lcc@{}}
\toprule
Model & Above teacher (\%)$\uparrow$ \rng{95\% CI} & Prompts W/T/L \\
\midrule
HY-Motion teacher @50 & 50 (reference) & -- \\
MotionLCM-V2 @4 & 41.2 \rng{30.4, 51.7} & 17/0/33 \\
MLCT @4 & 33.9 \rng{24.5, 43.4} & 11/0/39 \\
\rowcolor{methodwash}
\textbf{\tacd} @8 & \textbf{50.8} \rng{41.1, 60.5} & 28/1/21 \\
\bottomrule
\end{tabular}}
\caption{\textbf{Human preference:} win rate against the 50-step teacher and prompts won/tied/lost (\cref{app:human}).}
\label{tab:human}

\end{minipage}}%
\par\vspace{6pt}
{\renewcommand{\arraystretch}{1.15}%
\resizebox{\linewidth}{!}{%
\begin{tabular}{@{}lccc@{\hspace{4pt}}lr@{}}
\toprule
Configuration & Final-segment queries & $C$ & FID & \textcolor{tableink}{seed range} & vs.\ uncapped \\
\midrule
\rowcolor{amplifiedwash}
Uncapped student & $0.937,\ 0.999$ & $10^6$ & 2.486 & \rng{2.419--2.563} & -- \\
\rowcolor{methodwash}
\circnum{1}~\textbf{TACD} (grid capping) & $0.90625,\ 0.9375$ & $256$ & 1.041 & \rng{1.009--1.074} & \textcolor{tacdteal}{$-58\%$} \\
\circnum{2}~Endpoint matching & $0.937,\ 0.999$ & $1$ & 1.029 & \rng{0.988--1.091} & \textcolor{tacdteal}{$-59\%$} \\
\circnum{3}~Weight capping & $0.937,\ 0.999$ & $256$ & 1.011 & \rng{0.980--1.044} & \textcolor{tacdteal}{$-59\%$} \\
\quad\textit{Last-point-only} & $0.937,\ 0.9375$ & $256$ & 1.039 & \rng{1.032--1.048} & \textcolor{tacdteal}{$-58\%$} \\
\rowcolor{amplifiedwash}
\quad\textit{Capped grid, $C$ restored to $10^6$} & $0.90625,\ 0.9375$ & $10^6$ & 2.669 & \rng{2.645--2.703} & \textcolor{roseink}{$+7\%$} \\
\bottomrule
\end{tabular}}}
\caption{\textbf{Bounding the terminal coefficient $C$}, eight-step Lite: each bound (\circnum{1}--\circnum{3}) cuts FID by 58--59\%; restoring $C=10^6$ (rose) undoes it. Brackets: three-seed range.}
\label{tab:controls}
\label{tab:matched_training}

\end{minipage}\hfill
\usebox{\ssaebox}
\end{table}

\subsection{Few-step motion quality}
\label{sec:overall}
\label{sec:frontier}
On HumanML3D, at the same eight-step Lite budget, TACD reduces mean FID from $2.486$ to $1.041$ and improves R@3
from $.708$ to $.782$ (\cref{tab:matched_lite}); the uncapped student uses
$\pi$-Flow's single-knot DX policy on a fixed supervision grid (\cref{sec:degenerate}), so the comparison
evaluates terminal amplification control. The ports, our reimplementations of DMD~\citep{yin2024dmd}, consistency distillation~\citep{luo2023lcm}, and
Reflow~\citep{liu2023rectified} on the same Lite student, share its architecture, prompts,
and generator-update budget (\cref{sec:ports}); Reflow gives FID $3.341$ and R@3 $.660$, no better
than the teacher truncated to eight steps ($3.250$/$.652$). Under the MARDM evaluator
(\cref{tab:mardm}), \tacd\ leads the truncated teacher and every eight-step student in each seed and cuts the uncapped student's FID from $7.559$ to
$3.575$.
\begin{table}[t]
\scriptsize
\setlength{\tabcolsep}{3pt}
\renewcommand{\arraystretch}{1.0}
\begin{minipage}[c]{0.66\linewidth}
\centering
\resizebox{\linewidth}{!}{%
\begin{tabular}{@{}lcccccccc@{}}
\toprule
Method & Training signal & NFE & R@1$\uparrow$ & R@3$\uparrow$ & MM-Dist$\downarrow$ & FID$\downarrow$ & Div$\rightarrow$ & MMod$\uparrow$ \\
\midrule
\multicolumn{9}{@{}l}{\emph{MLD teacher}~\citep{chen2023mld} \emph{and MotionLCM-V1}~\citep{dai2024motionlcm}} \\
MLD teacher (DDIM) & Reference & 50 & .502 & .795 & 3.048 & .387 & 9.53 & -- \\
Truncated MLD teacher & None & 4 & .459 & .768 & 3.295 & .585 & 9.49 & -- \\
MotionLCM-V1 & Real motions & 4 & -- & .798 & -- & .286 & -- & -- \\
\rowcolor{methodwash}
\textbf{\tacd} & Teacher outputs & 4 & \textbf{.525} & \textbf{.818} & \textbf{2.909} & \textbf{.222} & 9.80 & -- \\
\midrule
\multicolumn{9}{@{}l}{\emph{MotionLCM-V2 teacher and student}~\citep{dai2024motionlcmv2}} \\
MotionLCM-V2 teacher & Reference & 50 & .537 & .824 & 2.838 & .066 & 9.59 & -- \\
MotionLCM-V2 student & Real motions & 4 & \textbf{.554} & \textbf{.841} & \textbf{2.760} & .072 & 9.54 & 1.72 \\
\rowcolor{methodwash}
\textbf{\tacd} & Teacher outputs & 4 & .538 & .824 & 2.838 & \textbf{.055} & 9.57 & \textbf{1.94} \\
\bottomrule
\end{tabular}}
\end{minipage}\hfill
\begin{minipage}[c]{\dimexpr0.97\linewidth-0.66\linewidth\relax}
\caption{\textbf{Diffusion teachers on HumanML3D.} Four-step \tacd\ has lower FID than its 50-step teacher and the released MotionLCM student (our evaluation); bold: best at four steps in each group.}
\label{tab:context}
\end{minipage}
\end{table}

\tacd\ (HY-Motion) exceeds the R@3 of
the 50-step teacher ($.743$) and of the released Lite at 50 steps ($.728$), while the teacher
retains better FID and diversity. On SSAE, a VLM-judged semantic-alignment score (\cref{tab:ssae}), \tacd\
scores highest ($.720$): $.020$--$.028$ above the 50-step teacher, the released
Lite, and the Consistency port and $.05$--$.13$ above the DMD, truncated-teacher, uncapped, and Reflow students. When every model, \tacd\ included ($.699$), is
rendered from joints on one skeleton, \tacd\ leads seven of ten open-weight HumanML3D models by $.03$--$.11$, including
the four-step MotionLCM and MLCT, and shows no detectable difference from MoGeFlow-L, MotionHiFlow, and
MotionLCM-V2. In a blind ranking study (\cref{tab:human}), annotators rank eight-step \tacd\ above its
50-step teacher in $51\%$ of comparisons, on par with it, while the four-step MotionLCM-V2 and MLCT reach
$41\%$ and $34\%$. For the teacher, the best FID and R@3 over CFG
$\{2,3,5,7.5\}$ both occur at 5, the value \tacd\ (HY-Motion) uses (\cref{tab:frontier}).

\subsection{Supervision analysis and controlled interventions}
\label{sec:onestate}
\label{sec:repair}
\label{sec:interventions}
\label{sec:algo}
At $K=8$ and $M=2$, the original supervision grid queries the final segment at $.937$ and $.999$:
the terminal coefficient is $10^6$, while the other fifteen coefficients are at most $252$. We first
measure where the gradient goes, then intervene on this one coefficient
(\cref{tab:controls,tab:gradient_cross}).

\paragraph{The final segment takes most of the gradient norm.}
We measure $\rho_k=\|\partial\mathcal L_k/\partial\theta\|/
\sum_r\|\partial\mathcal L_r/\partial\theta\|$ on the full Lite backbone before clipping.
At initialization, the original supervision grid assigns $68.3\%$ of the summed segment-gradient norms
to the final segment, versus $.88\%$ on the capped grid; the uncapped 8k checkpoint reaches $96.6\%$.
Probing both checkpoints on both grids (\cref{tab:gradient_cross}), the original grid puts
$96.6$--$97.6\%$ of the gradient norm on the final segment and the capped grid $3.6$--$11.8\%$, so
concentration follows the grid rather than the trained weights.

\paragraph{Bounding the coefficient restores quality; restoring it reverses the gain.}
All three enforcements of the bound of \cref{sec:tacd_repairs} recover FID to $1.01$--$1.04$
(\cref{tab:controls}; per-seed values in \cref{tab:coefficient_seeds}): grid capping (\circnum{1},
TACD), endpoint matching (\circnum{2}, the form used for diffusion teachers), and weight capping
(\circnum{3}), which sets the terminal coefficient to $C=4K^2=256$ on the original supervision grid
and leaves every other term unchanged (\cref{eq:intervention}). Conversely, restoring $C=10^6$ on the
capped grid, without moving any query, worsens FID in all three seeds, to a mean of $2.669$.

\paragraph{One query carries the effect.}
Keeping $.937$ and moving only $.999$ to $.9375$ gives FID $1.039$, close to full capping's $1.041$,
with improvement in all three seeds (with the 1B student at four steps: $3.291$ uncapped, $1.458$ fully capped, $1.462$ with
only the last point moved, \cref{app:diagrows}).

\paragraph{Relative weight, not total loss scale.}
The coefficient also changes the magnitude of the loss, but the optimizer removes that difference.
Gradient clipping at unit norm binds on at least 799 of the 800 logged updates in every run of
\cref{tab:controls}: the median pre-clip norm is $245$--$263$ for the uncapped student and $5.0$--$6.1$ for
the bounded students (\cref{tab:gnorm}). Clipping normalizes the gradient magnitude entering AdamW,
so the interventions act through the relative supervision weights (the terminal share of
\cref{tab:gradient_cross}).

\begin{table}[t]
\scriptsize
\begin{minipage}[t]{0.64\linewidth}
\vspace{0pt}
\centering
\includegraphics[width=\linewidth]{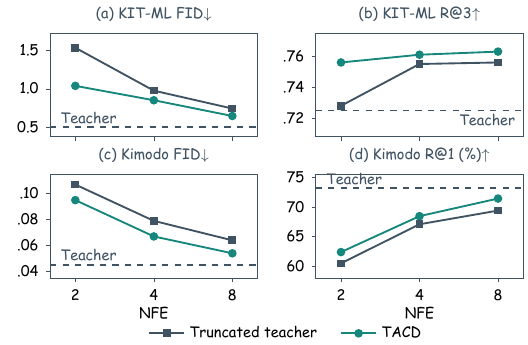}
\asfigure
\caption{\textbf{Few-step curves.} On KIT-ML (MDM teacher) and Kimodo, \tacd\ (teal) beats the truncated teacher (grey) at every NFE; dashed: the full-step teacher.}
\label{fig:fewstep}
\end{minipage}\hfill
\begin{minipage}[t]{0.34\linewidth}
\vspace{0pt}
\centering
\includegraphics[width=\linewidth]{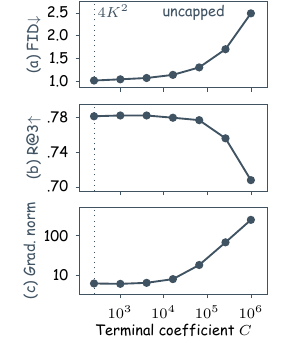}
\asfigure
\caption{\textbf{Sweeping $C$} on the eight-step Lite student: \tacd's $4K^{2}$ (dotted) sits on the plateau.}
\label{fig:coefsweep}
\end{minipage}
\end{table}

\paragraph{Sensitivity to the terminal coefficient.}
Sweeping the terminal coefficient (\cref{fig:coefsweep}; every run in \cref{app:coefsweep})
gives a dose--response: FID stays at $1.01$--$1.07$ from $C=256$ to $4096$ and rises monotonically beyond
it, to $1.14$, $1.30$, $1.70$, and $2.49$ at $C=16384$, $65536$, $262144$, and $10^6$, while the median
pre-clip norm rises in step from $6$ to $253$. The other metrics move with FID at every
step, and moving the grid instead ($\delta\in\{1/32,3/32\}$) stays in the plateau: $C=4K^2$ sits
in a plateau at least $16\times$ wide.

\begin{finding}
\textbf{Key finding.} In fixed-grid flow distillation, velocity matching can repeatedly
overweight errors near the denoising endpoint. TACD controls this amplification with a
step-size-aware bound, reducing eight-step HY-Motion student FID by \textbf{58\% versus
uncapped distillation} on HumanML3D without changing inference. Restoring the full weight
reverses the gain.
\end{finding}

\subsection{Other teachers and datasets}
\label{sec:extension}
\paragraph{Diffusion teachers on HumanML3D.}
\label{sec:closeddata}
With endpoint matching, four-step TACD improves on the 50-step MLD teacher in both FID
($.387\to.222$) and R@3 ($.795\to.818$), and its FID is below the $.286$ we measure for the
released MotionLCM-V1 checkpoint, a student trained on real motions (\cref{tab:context}). With the
MotionLCM-V2 teacher, four-step TACD reaches FID $.055$ against the teacher's $.066$ and the
released official student's $.072$ under the same evaluation, at the teacher's R@3 ($.824$). \tacd\ FID is the mean of three training seeds
($\pm.002$ for MLD, $\pm.001$ for MotionLCM-V2). Under the MARDM evaluator, four-step \tacd\ has FID $.228$
against the MLD teacher's $.512$ and matches real motions in R-precision (\cref{tab:mardm}).

\paragraph{KIT-ML.}
At the same NFE, TACD reduces FID from the truncated teacher's $1.530$, $.976$, and $.745$ to $1.038$,
$.852$, and $.649$ at two, four, and eight steps (\cref{fig:fewstep}a; \cref{tab:kit}b), $32\%$ lower at two
steps and $13\%$ at four and eight; our MDM teacher run agrees with its published row (\cref{tab:kit}a).

\paragraph{Kimodo.}
\label{sec:kimodo_main}
At eight steps, \tacd\ raises R@1 over the truncated teacher from $69.36\%$ to $71.39\%$ and lowers FID
from $.064$ to $.054$, with the same ordering at four and two steps (\cref{fig:fewstep}c,d; \cref{tab:kimodo_main}).

\subsection{Quality--efficiency trade-offs}
\label{sec:coupled}
\label{sec:crossscale}
\label{sec:modules}
\label{sec:quant}
\Cref{tab:efficiency,fig:cost} separate the savings from fewer sampling steps (computation) and
from smaller backbones or encoders (memory).
On HY-Motion, the Lite backbone and Qwen3-0.6B encoder of \tacd\ (HY-Motion) reduce
peak GPU memory from $17.63$ to $2.62$\,GB
($6.7\times$), mostly from the 8B-to-0.6B text encoder, and eight-step sampling cuts end-to-end
time on an RTX L40 (\texttt{bf16}, batch 1) from the teacher's $2077$ to $268$\,ms ($7.7\times$); the released Lite at 50 steps takes $1476$\,ms (\cref{tab:latency}).

\tacd\ (Kimodo), with a trained-in 1B encoder, reduces peak memory from $15.3$ to $4.0$\,GB ($3.8\times$) and
end-to-end time from $2.96$ to $.25$\,s ($11.9\times$) at the R@3 and FID of the 8B-encoder student
(\cref{tab:efficiency,tab:item7}).
\Cref{fig:teaser3} shows sample motions (\cref{app:teaser3}).

\section{Conclusion}
\label{sec:conclusion}
TACD trains efficient text-to-motion students from text prompts and pretrained teachers alone.
Where supervision is taken does not settle how strongly each error counts: on a fixed grid, velocity
matching overweights the near-terminal query at every update, and a step-size bound on that
weight restores few-step quality; compact encoders and backbones then cut latency and memory.

\FloatBarrier
\section*{AI use statement}
AI tools were used to polish manuscript wording and LaTeX formatting, assist with literature retrieval and discovery, and generate synthetic data used in our experiments.

\section*{Reproducibility statement}
The supplementary material provides complete flow pseudocode and objective definitions
(\cref{sec:exactobj}), training and evaluation settings (\cref{app:training_settings}),
Kimodo evaluation and compact-encoder comparisons (\cref{app:kimodo}), and
per-seed controls and gradient probes (\cref{app:diagrows}).

\bibliography{references}

\begin{thebibliography}{52}
\providecommand{\natexlab}[1]{#1}
\providecommand{\url}[1]{\texttt{#1}}
\expandafter\ifx\csname urlstyle\endcsname\relax
  \providecommand{\doi}[1]{doi: #1}\else
  \providecommand{\doi}{doi: \begingroup \urlstyle{rm}\Url}\fi

\bibitem[Amin et~al.(2025)Amin, Afroz, Gharami, Moghadampanah, and Nikolopoulos]{amin2025diffpro}
Farhana Amin, Sabiha Afroz, Kanchon Gharami, Mona Moghadampanah, and Dimitrios~S. Nikolopoulos.
\newblock {DiffPro}: Joint timestep and layer-wise precision optimization for efficient diffusion inference.
\newblock \emph{arXiv preprint arXiv:2511.11446}, 2025.

\bibitem[Chen et~al.(2026)Chen, Zhang, Tan, Guibas, Wetzstein, and Bi]{piflow2025}
Hansheng Chen, Kai Zhang, Hao Tan, Leonidas Guibas, Gordon Wetzstein, and Sai Bi.
\newblock pi-flow: Policy-based few-step generation via imitation distillation.
\newblock In \emph{ICLR}, 2026.
\newblock arXiv:2510.14974.

\bibitem[Chen et~al.(2023)Chen, Jiang, Liu, Huang, Fu, Chen, and Yu]{chen2023mld}
Xin Chen, Biao Jiang, Wen Liu, Zilong Huang, Bin Fu, Tao Chen, and Gang Yu.
\newblock Executing your commands via motion diffusion in latent space.
\newblock In \emph{CVPR}, 2023.

\bibitem[Dai et~al.(2024{\natexlab{a}})Dai, Chen, Huo, Wang, Liu, Dai, and Tang]{dai2024motionlcmv2}
Wenxun Dai, Ling-Hao Chen, Yufei Huo, Jingbo Wang, Jinpeng Liu, Bo~Dai, and Yansong Tang.
\newblock Real-time controllable motion generation via latent consistency model.
\newblock MotionLCM-V2 technical report, 2024{\natexlab{a}}.
\newblock \url{https://dai-wenxun.github.io/files/MotionLCM_V2.pdf}.

\bibitem[Dai et~al.(2024{\natexlab{b}})Dai, Chen, Wang, Liu, Dai, and Tang]{dai2024motionlcm}
Wenxun Dai, Ling-Hao Chen, Jingbo Wang, Jinpeng Liu, Bo~Dai, and Yansong Tang.
\newblock Motionlcm: Real-time controllable motion generation via latent consistency model.
\newblock In \emph{ECCV}, 2024{\natexlab{b}}.

\bibitem[Du et~al.(2026)Du, Jin, Xu, Gong, Gu, Yong, Guo, and Liu]{du2026wanquant}
Jinyang Du, Shenghao Jin, Ziqian Xu, Ruihao Gong, Shiqiao Gu, Yang Yong, Jinyang Guo, and Xianglong Liu.
\newblock Collaborative few-step distillation and low-bit quantization for {Wan2.2} dual-expert video diffusion models.
\newblock \emph{arXiv preprint arXiv:2606.00658}, 2026.

\bibitem[Fang et~al.(2026{\natexlab{a}})Fang, Sun, Zhan, Cai, and Fu]{moge_flow}
Pengcheng Fang, Tengjiao Sun, Xiaoyu Zhan, Xiaohao Cai, and Dongjie Fu.
\newblock {MoGeFlow}: Flowing through motion codebook geometry for text-to-motion generation.
\newblock arXiv preprint arXiv:2606.11656, 2026{\natexlab{a}}.

\bibitem[Fang et~al.(2026{\natexlab{b}})Fang, Huang, Zeng, Zhao, Chen, Feng, Lin, Chen, Chen, Cao, and Zhao]{flowopd2026}
Zhen Fang, Wenxuan Huang, Yu~Zeng, Yiming Zhao, Shuang Chen, Kaituo Feng, Yunlong Lin, Lin Chen, Zehui Chen, Shaosheng Cao, and Feng Zhao.
\newblock {Flow-OPD}: On-policy distillation for flow matching models.
\newblock \emph{arXiv preprint arXiv:2605.08063}, 2026{\natexlab{b}}.

\bibitem[Geng et~al.(2025)Geng, Deng, Bai, Kolter, and He]{geng2025meanflow}
Zhengyang Geng, Mingyang Deng, Xingjian Bai, J.~Zico Kolter, and Kaiming He.
\newblock Mean flows for one-step generative modeling.
\newblock \emph{arXiv:2505.13447}, 2025.

\bibitem[Gu et~al.(2023)Gu, Zhai, Zhang, Liu, and Susskind]{gu2023boot}
Jiatao Gu, Shuangfei Zhai, Yizhe Zhang, Lingjie Liu, and Joshua~M. Susskind.
\newblock Boot: Data-free distillation of denoising diffusion models with bootstrapping, 2023.
\newblock arXiv:2306.05544.

\bibitem[Gu et~al.(2026)Gu, Fang, Jiang, Mao, Han, Cai, and Shou]{anyflow2026}
Yuchao Gu, Guian Fang, Yuxin Jiang, Weijia Mao, Song Han, Han Cai, and Mike~Zheng Shou.
\newblock {AnyFlow}: Any-step video diffusion model with on-policy flow map distillation.
\newblock \emph{arXiv preprint arXiv:2605.13724}, 2026.

\bibitem[Guo et~al.(2022)Guo, Zou, Zuo, Wang, Ji, Li, and Cheng]{guo2022humanml3d}
Chuan Guo, Shihao Zou, Xinxin Zuo, Sen Wang, Wei Ji, Xingyu Li, and Li~Cheng.
\newblock Generating diverse and natural 3d human motions from text.
\newblock In \emph{CVPR}, 2022.

\bibitem[Guo et~al.(2024)Guo, Mu, Javed, Wang, and Cheng]{guo2024momask}
Chuan Guo, Yuxuan Mu, Muhammad~Gohar Javed, Sen Wang, and Li~Cheng.
\newblock Momask: Generative masked modeling of 3d human motions.
\newblock In \emph{CVPR}, 2024.

\bibitem[Hang et~al.(2023)Hang, Gu, Li, Bao, Chen, Hu, Geng, and Guo]{minsnr2023}
Tiankai Hang, Shuyang Gu, Chen Li, Jianmin Bao, Dong Chen, Han Hu, Xin Geng, and Baining Guo.
\newblock Efficient diffusion training via min-snr weighting strategy.
\newblock In \emph{ICCV}, 2023.

\bibitem[He et~al.(2024)He, Liu, Wu, Zhou, and Zhuang]{he2024efficientdm}
Yefei He, Jing Liu, Weijia Wu, Hong Zhou, and Bohan Zhuang.
\newblock {EfficientDM}: Efficient quantization-aware fine-tuning of low-bit diffusion models.
\newblock In \emph{International Conference on Learning Representations (ICLR)}, 2024.

\bibitem[Hong et~al.(2026)Hong, Liu, Bian, Wang, and Zhang]{hong2026alignment}
Jiadong Hong, Lei Liu, Xinyu Bian, Wenjie Wang, and Zhaoyang Zhang.
\newblock Prediction--loss alignment for sampler--robust flow matching training.
\newblock \emph{arXiv preprint arXiv:2602.10420}, 2026.

\bibitem[Hu et~al.(2024)Hu, Zhu, Zhou, Yan, Li, Liu, and Chen]{hu2024mlct}
Mengxian Hu, Minghao Zhu, Xun Zhou, Qingqing Yan, Shu Li, Chengju Liu, and Qijun Chen.
\newblock Efficient text-driven motion generation via latent consistency training.
\newblock \emph{arXiv preprint arXiv:2405.02791}, 2024.

\bibitem[Jiang et~al.(2023)Jiang, Chen, Liu, Yu, Yu, and Chen]{jiang2023motiongpt}
Biao Jiang, Xin Chen, Wen Liu, Jingyi Yu, Gang Yu, and Tao Chen.
\newblock Motiongpt: Human motion as a foreign language.
\newblock In \emph{NeurIPS}, 2023.

\bibitem[Jiang et~al.(2025)Jiang, Wei, and Ni]{motionpcm2025}
Lei Jiang, Ye~Wei, and Hao Ni.
\newblock Motionpcm: Real-time motion synthesis with phased consistency model.
\newblock \emph{arXiv preprint arXiv:2501.19083}, 2025.

\bibitem[Karras et~al.(2022)Karras, Aittala, Aila, and Laine]{karras2022edm}
Tero Karras, Miika Aittala, Timo Aila, and Samuli Laine.
\newblock Elucidating the design space of diffusion-based generative models.
\newblock In \emph{NeurIPS}, 2022.

\bibitem[Li et~al.(2026{\natexlab{a}})Li, Lin, Zeng, Kang, Li, and Hu]{motionhiflow}
Heng Li, Xiaotong Lin, Ling-An Zeng, Yulei Kang, Shuai Li, and Jian-Fang Hu.
\newblock {MotionHiFlow}: Text-to-motion via hierarchical flow matching.
\newblock In \emph{CVPR}, 2026{\natexlab{a}}.
\newblock arXiv:2604.23264.

\bibitem[Li \& He(2025)Li and He]{li2025jit}
Tianhong Li and Kaiming He.
\newblock Back to basics: Let denoising generative models denoise.
\newblock \emph{arXiv preprint arXiv:2511.13720}, 2025.

\bibitem[Li et~al.(2023)Li, Liu, Lian, Yang, Dong, Kang, Zhang, and Keutzer]{li2023qdiffusion}
Xiuyu Li, Yijiang Liu, Long Lian, Huanrui Yang, Zhen Dong, Daniel Kang, Shanghang Zhang, and Kurt Keutzer.
\newblock {Q-Diffusion}: Quantizing diffusion models.
\newblock In \emph{IEEE/CVF International Conference on Computer Vision (ICCV)}, 2023.

\bibitem[Li et~al.(2026{\natexlab{b}})Li, Yang, Lei, Fu, Zeng, Hu, Wei, and Zheng]{irg_motionllm}
Yuan-Ming Li, Qize Yang, Nan Lei, Shenghao Fu, Ling-An Zeng, Jian-Fang Hu, Xihan Wei, and Wei-Shi Zheng.
\newblock {IRG-MotionLLM}: Interleaving motion generation, assessment and refinement for text-to-motion generation.
\newblock In \emph{ECCV}, 2026{\natexlab{b}}.
\newblock arXiv:2512.10730.

\bibitem[Lin et~al.(2024)Lin, Liu, Li, and Yang]{lin2024common}
Shanchuan Lin, Bingchen Liu, Jiashi Li, and Xiao Yang.
\newblock Common diffusion noise schedules and sample steps are flawed.
\newblock In \emph{WACV}, 2024.

\bibitem[Lipman et~al.(2023)Lipman, Chen, Ben-Hamu, Nickel, and Le]{lipman2023flowmatching}
Yaron Lipman, Ricky T.~Q. Chen, Heli Ben-Hamu, Maximilian Nickel, and Matthew Le.
\newblock Flow matching for generative modeling.
\newblock In \emph{ICLR}, 2023.

\bibitem[Liu et~al.(2023)Liu, Gong, and Liu]{liu2023rectified}
Xingchao Liu, Chengyue Gong, and Qiang Liu.
\newblock Flow straight and fast: Learning to generate and transfer data with rectified flow.
\newblock In \emph{ICLR}, 2023.

\bibitem[Liu et~al.(2024)Liu, Zhang, Ma, Peng, and Liu]{liu2024instaflow}
Xingchao Liu, Xiwen Zhang, Jianzhu Ma, Jian Peng, and Qiang Liu.
\newblock Instaflow: One step is enough for high-quality diffusion-based text-to-image generation.
\newblock In \emph{ICLR}, 2024.

\bibitem[Lu et~al.(2022)Lu, Zhou, Bao, Chen, Li, and Zhu]{lu2022dpmsolver}
Cheng Lu, Yuhao Zhou, Fan Bao, Jianfei Chen, Chongxuan Li, and Jun Zhu.
\newblock {DPM-Solver}: A fast {ODE} solver for diffusion probabilistic model sampling in around 10 steps.
\newblock In \emph{Advances in Neural Information Processing Systems (NeurIPS)}, 2022.

\bibitem[Luo et~al.(2023)Luo, Tan, Huang, Li, and Zhao]{luo2023lcm}
Simian Luo, Yiqin Tan, Longbo Huang, Jian Li, and Hang Zhao.
\newblock Latent consistency models: Synthesizing high-resolution images with few-step inference.
\newblock \emph{arXiv:2310.04378}, 2023.

\bibitem[Luo et~al.(2026)Luo, Qian, Tang, Zhu, Liu, Wang, and You]{afd2026}
Yang Luo, Shengju Qian, Xiaohang Tang, Zirui Zhu, Yong Liu, Xin Wang, and Yang You.
\newblock On-policy adversarial flow distillation for autoregressive video generation.
\newblock \emph{arXiv preprint arXiv:2605.26105}, 2026.

\bibitem[Luo et~al.(2025)Luo, Hu, Sun, Cai, and Tang]{luo2025tdm}
Yihong Luo, Tianyang Hu, Jiacheng Sun, Yujun Cai, and Jing Tang.
\newblock Learning few-step diffusion models by trajectory distribution matching.
\newblock \emph{arXiv:2503.06674}, 2025.

\bibitem[Meng et~al.(2025{\natexlab{a}})Meng, Han, Peng, Xie, and Jiang]{meng2025acmdm}
Zichong Meng, Zeyu Han, Xiaogang Peng, Yiming Xie, and Huaizu Jiang.
\newblock Absolute coordinates make motion generation easy.
\newblock \emph{arXiv preprint arXiv:2505.19377}, 2025{\natexlab{a}}.

\bibitem[Meng et~al.(2025{\natexlab{b}})Meng, Xie, Peng, Han, and Jiang]{meng2025mardm}
Zichong Meng, Yiming Xie, Xiaogang Peng, Zeyu Han, and Huaizu Jiang.
\newblock Rethinking diffusion for text-driven human motion generation: Redundant representations, evaluation, and masked autoregression.
\newblock In \emph{CVPR}, 2025{\natexlab{b}}.

\bibitem[Petrovich et~al.(2022)Petrovich, Black, and Varol]{petrovich2022temos}
Mathis Petrovich, Michael~J. Black, and G{\"u}l Varol.
\newblock Temos: Generating diverse human motions from textual descriptions.
\newblock In \emph{ECCV}, 2022.

\bibitem[Petrovich et~al.(2023)Petrovich, Black, and Varol]{petrovich2023tmr}
Mathis Petrovich, Michael~J. Black, and G{\"u}l Varol.
\newblock {TMR}: Text-to-motion retrieval using contrastive {3D} human motion synthesis.
\newblock In \emph{ICCV}, 2023.

\bibitem[Plappert et~al.(2016)Plappert, Mandery, and Asfour]{plappert2016kit}
Matthias Plappert, Christian Mandery, and Tamim Asfour.
\newblock The {KIT} motion-language dataset.
\newblock \emph{Big Data}, 4\penalty0 (4):\penalty0 236--252, 2016.

\bibitem[Rempe et~al.(2026)Rempe, Petrovich, Yuan, Zhang, Peng, Jiang, Wang, Iqbal, Minor, de~Ruyter, Li, Tessler, Lim, Jeong, Wu, Hassani, Huang, Yu, Chung, Song, Dionne, Kautz, Yuen, and Fidler]{rempe2026kimodo}
Davis Rempe, Mathis Petrovich, Ye~Yuan, Haotian Zhang, Xue~Bin Peng, Yifeng Jiang, Tingwu Wang, Umar Iqbal, David Minor, Michael de~Ruyter, Jiefeng Li, Chen Tessler, Edy Lim, Eugene Jeong, Sam Wu, Ehsan Hassani, Michael Huang, Jin-Bey Yu, Chaeyeon Chung, Lina Song, Olivier Dionne, Jan Kautz, Simon Yuen, and Sanja Fidler.
\newblock {Kimodo}: Scaling controllable human motion generation.
\newblock \emph{arXiv preprint arXiv:2603.15546}, 2026.

\bibitem[Salimans \& Ho(2022)Salimans and Ho]{salimans2022progressive}
Tim Salimans and Jonathan Ho.
\newblock Progressive distillation for fast sampling of diffusion models.
\newblock In \emph{ICLR}, 2022.

\bibitem[Salimans et~al.(2024)Salimans, Mensink, Heek, and Hoogeboom]{salimans2024moment}
Tim Salimans, Thomas Mensink, Jonathan Heek, and Emiel Hoogeboom.
\newblock Multistep distillation of diffusion models via moment matching.
\newblock \emph{arXiv:2406.04103}, 2024.

\bibitem[Song et~al.(2023)Song, Dhariwal, Chen, and Sutskever]{song2023consistency}
Yang Song, Prafulla Dhariwal, Mark Chen, and Ilya Sutskever.
\newblock Consistency models.
\newblock In \emph{ICML}, 2023.

\bibitem[Sun et~al.(2024)Sun, Tang, Wang, Meng, Jiang, Ma, and Zhu]{sun2024tmpqdm}
Haojun Sun, Chen Tang, Zhi Wang, Yuan Meng, Jingyan Jiang, Xinzhu Ma, and Wenwu Zhu.
\newblock {TMPQ-DM}: Joint timestep reduction and quantization precision selection for efficient diffusion models.
\newblock \emph{arXiv preprint arXiv:2404.09532}, 2024.

\bibitem[Tevet et~al.(2023)Tevet, Raab, Gordon, Shafir, Cohen-Or, and Bermano]{tevet2023mdm}
Guy Tevet, Sigal Raab, Brian Gordon, Yonatan Shafir, Daniel Cohen-Or, and Amit~H. Bermano.
\newblock Human motion diffusion model.
\newblock In \emph{ICLR}, 2023.

\bibitem[Wen et~al.(2025)Wen, Shuai, Kang, Li, Wen, Qian, Jiao, Chen, Chen, Wang, Guo, An, Liu, Tong, Zhang, Guo, Chen, Zhang, Zhang, Yao, Zhang, Duan, Wu, Chen, Cheng, Dong, He, Zhang, Lin, Zhang, Fan, Li, Hu, Liu, {Linus}, Jiang, Li, and Bao]{hymotion2025}
Yuxin Wen, Qing Shuai, Di~Kang, Jing Li, Cheng Wen, Yue Qian, Ningxin Jiao, Changhai Chen, Weijie Chen, Yiran Wang, Jinkun Guo, Dongyue An, Han Liu, Yanyu Tong, Chao Zhang, Qing Guo, Juan Chen, Qiao Zhang, Youyi Zhang, Zihao Yao, Cheng Zhang, Hong Duan, Xiaoping Wu, Qi~Chen, Fei Cheng, Liang Dong, Peng He, Hao Zhang, Jiaxin Lin, Chao Zhang, Zhongyi Fan, Yifan Li, Zhichao Hu, Yuhong Liu, {Linus}, Jie Jiang, Xiaolong Li, and Linchao Bao.
\newblock Hy-motion 1.0: Scaling flow matching models for text-to-motion generation.
\newblock \emph{arXiv:2512.23464}, 2025.

\bibitem[Xiang et~al.(2025)Xiang, Zhang, Shang, Wu, Yan, and Nie]{xiang2025dkdm}
Qianlong Xiang, Miao Zhang, Yuzhang Shang, Jianlong Wu, Yan Yan, and Liqiang Nie.
\newblock {DKDM}: Data-free knowledge distillation for diffusion models with any architecture.
\newblock In \emph{Proceedings of the IEEE/CVF Conference on Computer Vision and Pattern Recognition (CVPR)}, 2025.

\bibitem[Ye et~al.(2026)Ye, Ge, Zhang, Li, Zhu, Zhao, Liu, Wang, Zhu, Yu, and Xiao]{flowcts2026}
Kaiyang Ye, Yuan Ge, Junxiang Zhang, Bei Li, Ziming Zhu, Haishu Zhao, Xiaoqian Liu, Chenglong Wang, Jingbo Zhu, Zhengtao Yu, and Tong Xiao.
\newblock Flowcts: On-policy continuous trajectory supervision of flow models.
\newblock \emph{arXiv preprint arXiv:2607.24522}, 2026.

\bibitem[Yin et~al.(2024{\natexlab{a}})Yin, Gharbi, Park, Zhang, Shechtman, Durand, and Freeman]{yin2024dmd2}
Tianwei Yin, Micha{\"e}l Gharbi, Taesung Park, Richard Zhang, Eli Shechtman, Fr{\'e}do Durand, and William~T. Freeman.
\newblock Improved distribution matching distillation for fast image synthesis.
\newblock In \emph{NeurIPS}, 2024{\natexlab{a}}.

\bibitem[Yin et~al.(2024{\natexlab{b}})Yin, Gharbi, Zhang, Shechtman, Durand, Freeman, and Park]{yin2024dmd}
Tianwei Yin, Micha{\"e}l Gharbi, Richard Zhang, Eli Shechtman, Fr{\'e}do Durand, William~T. Freeman, and Taesung Park.
\newblock One-step diffusion with distribution matching distillation.
\newblock In \emph{CVPR}, 2024{\natexlab{b}}.

\bibitem[Zhang et~al.(2023{\natexlab{a}})Zhang, Zhang, Cun, Zhang, Zhao, Lu, Shen, and Shan]{guo2022t2mgpt}
Jianrong Zhang, Yangsong Zhang, Xiaodong Cun, Yong Zhang, Hongwei Zhao, Hongtao Lu, Xi~Shen, and Ying Shan.
\newblock Generating human motion from textual descriptions with discrete representations.
\newblock In \emph{CVPR}, 2023{\natexlab{a}}.

\bibitem[Zhang et~al.(2023{\natexlab{b}})Zhang, Guo, Pan, Cai, Hong, Li, Yang, and Liu]{zhang2023remodiffuse}
Mingyuan Zhang, Xinying Guo, Liang Pan, Zhongang Cai, Fangzhou Hong, Huirong Li, Lei Yang, and Ziwei Liu.
\newblock Remodiffuse: Retrieval-augmented motion diffusion model.
\newblock In \emph{ICCV}, 2023{\natexlab{b}}.

\bibitem[Zhang et~al.(2024)Zhang, Cai, Pan, Hong, Guo, Yang, and Liu]{zhang2023motiondiffuse}
Mingyuan Zhang, Zhongang Cai, Liang Pan, Fangzhou Hong, Xinying Guo, Lei Yang, and Ziwei Liu.
\newblock Motiondiffuse: Text-driven human motion generation with diffusion model.
\newblock \emph{IEEE TPAMI}, 2024.

\bibitem[Zhou et~al.(2025)Zhou, Wang, Zheng, and Huang]{zhou2024sidlsg}
Mingyuan Zhou, Zhendong Wang, Huangjie Zheng, and Hai Huang.
\newblock Guided score identity distillation for data-free one-step text-to-image generation.
\newblock In \emph{International Conference on Learning Representations (ICLR)}, 2025.

\end{thebibliography}

\clearpage
\raggedbottom
\appendix
\crefalias{section}{appendix}      %
\crefalias{subsection}{appendix}
\section*{Supplementary material}
\startcontents[supp]
{\flushbottom\parskip=0pt plus 1pt\hypersetup{linkcolor=black}%
\printcontents[supp]{s}{1}{\setcounter{tocdepth}{2}}\par\penalty-10000}%
\clearpage\raggedbottom
\suppgroup{Method}
\section{Objective and implementation details}
\label{sec:exactobj}
This appendix expands \cref{sec:method} into what the trainer computes: the objectives that
\cref{sec:method} uses without displaying (\cref{app:objdefs}), the flow training iteration and its conventions
(\cref{app:conventions}), and the diffusion-teacher branch (\cref{app:diffbranch}).
\subsection{Objective definitions}
\label{app:objdefs}
The objects below enter training in this order: the rollout prediction, the gradient form of the
conversion factor, and the two loss variants used in the experiments.

\paragraph{Rollout prediction.} The EMA student $\bar\theta$ generates the rollout with the same
one-step parameterization as the online student (\cref{eq:fast_endpoint}). At each rollout segment
start $\bar x^{(k)}$ it predicts
\begin{equation}
\hat x_{1,\bar\theta}^{(k)}
=\bar x^{(k)}+(1-t_k)v_{\bar\theta}(\bar x^{(k)},t_k;c),
\label{eq:ema_endpoint}
\end{equation}
without gradient. This prediction fixes the next segment start through \cref{eq:inference_update} and
the supervision states through \cref{eq:ema_state}.

\paragraph{Gradient form of the conversion.} The supervision state and the teacher target are both
detached, so the factor $(1-s)^{-2}$ of \cref{eq:conversion} does not depend on $\theta$ and the
conversion holds for gradients, query by query:
\begin{equation}
\nabla_\theta\ell_v(t)=\frac{1}{(1-t)^2}\nabla_\theta\ell_x(t).
\label{eq:gradient_conversion}
\end{equation}
A query's gradient is its clean-motion gradient scaled by $(1-t)^{-2}$. How much the scaled query
contributes to an update therefore also depends on the size of $\nabla_\theta\ell_x$, which
\cref{sec:onestate} measures.

\paragraph{Endpoint matching.} Setting every coefficient $b_{k,j}$ of \cref{eq:unified} to one gives
\begin{equation}
\widetilde{\mathcal L}_{x}(\theta;\bar\theta)
=\sum_{k=0}^{K-1}\frac{w_k}{M}\sum_{j=1}^{M}
\left\|
\hat{x}_{1,\theta}^{(k)}-
\operatorname{sg}\!\left[
\hat x_1^{\mathrm{tea}}(\bar x^{(k,j)},s_{k,j};c)
\right]
\right\|_m^2 ,
\label{eq:endpoint}
\end{equation}
which compares the student's clean-motion prediction directly with the teacher's clean-motion
estimates at the supervision states. It is the third enforcement of the bound in
\cref{sec:tacd_repairs} and the form TACD takes for diffusion teachers (\cref{sec:ddim_instance}).

\paragraph{Coefficient interventions.} The interventions of \cref{sec:repair} and \cref{app:coefsweep}
change the weight of a single query. A weight $a_{k,j}$ on the velocity loss of query $(k,j)$ acts as
the endpoint coefficient
\begin{equation}
a_{k,j}\ell_v(s_{k,j})
=\underbrace{\frac{a_{k,j}}{(1-s_{k,j})^2}}_{\text{endpoint-equivalent coefficient}}
\ell_x(s_{k,j}),
\label{eq:intervention}
\end{equation}
so $a_{k,j}=C(1-s_{k,j})^2$ gives that query the coefficient $C$ while every other term and the
normalization $w_k/M$ stay unchanged. Weight capping is the case $C=4K^2$ on the terminal query.

\subsection{Training algorithm and conventions}
\label{app:conventions}
\Cref{alg:dm} gives one training iteration of the flow branch: \tacd's grid capping, or the
endpoint-matching control of \cref{sec:repair}. The paragraphs after it state the conventions the iteration relies on:
fixed supervision times, the endpoint form of the residual, and the rollout EMA.

\begin{algorithm}[b]%
\small
\SetAlgoLined
\DontPrintSemicolon
\KwIn{frozen teacher $\teacher$; online student $\theta$; rollout EMA $\bar\theta$;
boundaries $\{t_k\}_{k=0}^{K}$; control $q\in\{\mathrm{grid},\mathrm{endpoint}\}$}
Choose $\delta\leftarrow1/(2K)$ if $q=\mathrm{grid}$, else $\delta\leftarrow10^{-3}$;
construct $\{s_{k,j}(\delta)\}$ with \cref{eq:cap}\;
sample noise $\bar x^{(0)}$ and text $c$\;
\tcp{detached rollout-EMA trajectory}
\For{$k=0,\ldots,K-1$}{
  compute $\hat x_{1,\bar\theta}^{(k)}$ with \cref{eq:ema_endpoint}\;
  store $\bar x^{(k,j)}$ with \cref{eq:ema_state} for $j=1,\ldots,M$\;
  apply the update in \cref{eq:inference_update} with
  $\hat x_{1,\bar\theta}^{(k)}$ to obtain $\bar x^{(k+1)}$\;
}
$\mathcal L\leftarrow0$\;
\tcp{online prediction and frozen-teacher queries}
\For{$k=0,\ldots,K-1$}{
  compute $\hat x_{1,\theta}^{(k)}$ at detached $\bar x^{(k)}$\;
  \For{$j=1,\ldots,M$}{
    query $\hat x_1^{\rm tea}(\bar x^{(k,j)},s_{k,j};c)$ with \cref{eq:teacher_endpoint}\;
    set $\ell_q^{(k,j)}\leftarrow\ell_x(s_{k,j})$ if $q=\mathrm{endpoint}$,
    else $\ell_v(s_{k,j})$, using the pointwise residuals in \cref{eq:conversion}\;
    $\mathcal L\leftarrow\mathcal L+\frac{w_k}{M}\ell_q^{(k,j)}$\;
  }
}
update $\theta$; update rollout EMA $\bar\theta$ and the separate evaluation EMA\;
\caption{One training iteration of the flow branch. \tacd\ uses $q=\mathrm{grid}$ (capped supervision
grid, velocity loss); $q=\mathrm{endpoint}$ is the endpoint-matching control (original supervision grid,
endpoint loss). Only the online student receives gradients.}
\label{alg:dm}
\end{algorithm}

\paragraph{Supervision times are a fixed grid, not samples.} The loss of \cref{eq:lagged}
and the diffusion objective in \cref{eq:onx0} are computed with $M{=}2$ deterministic supervision
points per segment, $s_{k,j}=t_k+(s^{\mathrm{hi}}_k-t_k)\tfrac{j}{M}$ for $j{=}1,2$, so no randomness
enters beyond the noise $\epsilon$ and the prompt. For the flow case
$s^{\mathrm{hi}}_k=\min(t_{k+1},1-\delta)$: the policy velocity
$\pi_\theta(x_t,t)=(\xhat^{(k)}-x_t)/(1-t)$ is singular as $t\to1$, and the last segment's upper
endpoint is excluded by the margin $\delta$. The uncapped student
uses $\delta=10^{-3}$; TACD uses $\delta=1/(2K)$ on a uniform sampling grid (\cref{tab:configs}).
The last-point-only control moves only the last supervision time and leaves the other original
point unchanged.

\paragraph{The flow residual is the endpoint residual, reweighted.} \Cref{eq:conversion} holds
pointwise and for the same norm, so \cref{eq:lagged} is an endpoint-matching loss carrying an implicit
$(1-t)^{-2}$ weight. The flow endpoint-matching control removes this factor on the same states
(\cref{sec:repair}); \cref{eq:onx0} applies endpoint matching on the DDIM trajectory.
The norm is a mask-averaged squared error over valid frames and channels, with a factor $\tfrac12$
in the flow branch.

\paragraph{Rollout EMA.}
\label{app:stab}
\label{app:ema}
\label{sec:ema}
The rollout that fixes the segment starts $\bar x^{(k)}$ and the supervision states $\bar x^{(k,j)}$
comes from the EMA weights $\bar\theta$ and is detached; the frozen teacher supplies the targets, and
gradients pass only through the online student's segment-start endpoints (\cref{eq:lagged}). The diffusion
objective in \cref{eq:onx0} uses the same detached-state convention on its DDIM trajectory.
The rollout-EMA decay is $\mu{=}0.999$ on the flow teacher and $\mu{=}0.98$ on Kimodo; the evaluation
EMA is $0.999$ everywhere.

\subsection{Diffusion-teacher branch}
\label{app:diffbranch}
This subsection expands \cref{sec:ddim_instance}. For the $\epsilon$- and $x_0$-prediction diffusion
teachers, each segment is a deterministic ($\eta{=}0$) DDIM sub-trajectory, with $M$ supervision timesteps
$\tau_{k,j}=\tau_{k+1}+\frac{j}{M}{(\tau_k-\tau_{k+1})}$, $j=1,\ldots,M$, and $\tau_K=0$; the last is the segment start. The student commits to
one $\hat x_0^{(k)}$ per segment from a single gradient-carrying evaluation at the segment start, and
the supervision states lie on the \emph{detached} DDIM sub-trajectory induced by that constant
$\hat x_0^{(k)}$, built with the noise implied by the rollout prediction at the segment start,
\begin{equation}
\bar\epsilon^{(k)}=\frac{\bar x^{(k)}-\sqrt{\bar\alpha_{\tau_k}}\,\hat x_{0,\bar\theta}^{(k)}}
{\sqrt{1-\bar\alpha_{\tau_k}}}.
\label{eq:implied_noise}
\end{equation}
The student matches the teacher's per-step $\hat x_0$ at those states,
\begin{equation}
\mathcal L_{\mathrm{on}}^{x_0}
=\sum_k\frac{w_k}{M}\sum_j
\left\|\hat x_{0,\theta}^{(k)}
-\operatorname{sg}[\hat x_0^{\mathrm{tea}}(\bar x^{(k,j)},\tau_{k,j};c)]\right\|_m^2 ,
\label{eq:onx0}
\end{equation}
which is \cref{eq:unified} with $b_{k,j}\equiv1$; gradients pass only through the segment-start
predictions. Matching noise predictions on the same states would instead weight the clean-motion
error by the signal-to-noise ratio,
\begin{equation}
\left\|\hat\epsilon_\theta-\hat\epsilon^{\mathrm{tea}}\right\|_m^2
=\frac{\bar\alpha_\tau}{1-\bar\alpha_\tau}\,
\left\|\hat x_{0,\theta}^{(k)}-\hat x_0^{\mathrm{tea}}\right\|_m^2,
\label{eq:snr}
\end{equation}
which grows without bound toward the clean end; \cref{eq:onx0} carries no such weight.

\paragraph{Time direction.} Flow time runs $t_0{=}0\to t_K{=}1$ from noise to data, so segment $k$ is
$[t_k,t_{k+1}]$ with $t_k{<}t_{k+1}$. DDIM timesteps run the other way, $\tau_0{>}\tau_1{>}\dots$, so the same
segment is written $[\tau_{k+1},\tau_k]$ in \cref{eq:ddim_state,eq:onx0}. Both denote the same object within the
$k$-th sampler step.

\paragraph{Schedule and guidance in the diffusion branch.} $\bar\alpha_\tau$ at a continuous $\tau$ is a
linear interpolation of the teacher's own \texttt{alphas\_cumprod} between $\lfloor\tau\rfloor$ and
$\lceil\tau\rceil$; teacher and student share that array, the same $K$-step timestep schedule, and the same
classifier-free guidance scale applied identically to both (a guided evaluation counts once toward NFE,
\cref{app:evaluation}).

\FloatBarrier %
\section{Relation to the single-knot \texorpdfstring{$\pi$}{pi}-Flow objective}
\label{sec:degenerate}
The flow objective of \cref{sec:setting} uses the $N{=}1$ single-knot DX policy of
$\pi$-Flow~\citep{piflow2025}, written in our time convention, on a fixed supervision grid rather than
$\pi$-Flow's uniformly sampled query times.

\paragraph{Single-knot window identity.}
\label{lem:window}
Along the rollout induced by one fixed endpoint, the policy velocity $(\xhat-x_t)/(1-t)$ is constant, so
$\pi$-Flow's micro-window averaging (the policy velocity averaged over $[t,t{+}\Delta t]$) reduces to the
instantaneous velocity.

\paragraph{Numerical check.}
\label{sec:equiv}
On random \texttt{float64} states, the trainer's segment loss and its gradient with respect to the student's
endpoint agree with an independent implementation of the $N{=}1$ DX segment loss to $10^{-9}$ relative
tolerance, and the window average computed by quadrature matches the instantaneous value to the same
tolerance. The uncapped student in \cref{tab:matched_lite} therefore uses $\pi$-Flow's single-knot DX policy on
the fixed supervision grid.

\FloatBarrier %
\suppgroup{Experimental setup}
\section{Experimental settings}
\label{app:training_settings}
This appendix collects the settings shared by the experiments of \cref{sec:exp}: models and training
for each benchmark (\cref{app:models}), the evaluation protocol and model selection
(\cref{app:evaluation}), and the named training configurations (\cref{app:configs}).
\subsection{Models and training}
\label{app:models}
We evaluate few-step distillation on HumanML3D~\citep{guo2022humanml3d} and
KIT-ML~\citep{plappert2016kit}, using HY-Motion flow and MDM diffusion teachers, respectively.
Kimodo provides a second core system evaluation of few-step generation and compact conditioning
under a TMR-based protocol.
The controlled HumanML3D comparison uses HY-Motion's frozen 1B flow teacher with its native Qwen3-8B text encoder.
The student combines the 0.46B Lite motion backbone with a frozen Qwen3-0.6B encoder and a trainable
conditioning projection. The uncapped student, TACD, and the two other enforcements of its bound share initialization, a 6,000-prompt cache,
8,000 training iterations, learning rate $3\times10^{-6}$, effective batch size 4, and evaluation
EMA selection at the final iteration. All use uniform eight-step inference and CFG $5.0$.
They differ only in the supervision grid, matching space, or terminal coefficient defined in
\cref{sec:tacd_repairs,sec:interventions}.
Configurations and port-specific settings are in \cref{tab:configs,sec:ports}.

On KIT-ML, the teacher is the released 1000-step MDM~\citep{tevet2023mdm}, which predicts $x_0$
in a 251-d raw-motion representation. We train teacher-initialized, rank-32 LoRA students at
two, four, and eight steps with the DDIM endpoint-matching recipe in \cref{sec:ddim_instance}, using
trailing timestep spacing~\citep{lin2024common} for the student and the truncated teacher alike.
Training uses 4,000 text prompts, batch size 32, learning rate $10^{-4}$, CFG $2.5$, rollout EMA,
and 10,000 iterations. Our students on both benchmarks train with teacher supervision and no real motions.

For Kimodo~\citep{rempe2026kimodo}, the teacher is a released 100-step $x_0$-diffusion model over a 77-joint skeleton.
The student retains the teacher's 0.28B motion backbone and uses low-rank
adaptation. \tacd\ (Kimodo) conditions on a Llama-3.2-1B encoder through a layer-fusion
head and motion adapters.
We use DDIM endpoint matching with the teacher's 8B text encoder at $K\in\{8,4,2\}$.
Evaluation uses 607 prompts in a TMR-based harness.
Of these, 567 have ground-truth motions, used for evaluation only; stratum results and training settings are in \cref{app:kimodo}.
Generation latency is measured end-to-end for eight-second motions on one RTX L40 (\cref{sec:latency}).
All 8B-encoder rows have 15.3\,GB measured peak memory.
In the trained-in encoder comparison, both encoders train with global batch 64 at $K=8$ (the 8B
encoder split across 2--4 GPUs by data parallelism, drawing the same global minibatch). Complete paired records are in \cref{tab:item7}.

\subsection{Evaluation}
\label{app:evaluation}
HumanML3D and KIT-ML use the dataset-specific evaluators of \citet{guo2022humanml3d}: a text encoder and a
motion encoder trained so that matching text--motion pairs lie close in a shared feature space. The quality,
alignment, and diversity metrics below are computed on these features, with the real test motions as the
reference, and averaged over 20 evaluator repetitions; HumanML3D uses rerank$_k=1$ and \texttt{fp32} evaluation. \Cref{app:mardm} repeats the
HumanML3D comparisons with the 67-dimensional evaluators of \citet{meng2025mardm}. In the tables,
$\uparrow$ and $\downarrow$ mark the better direction, and $\rightarrow$ means closer to real motions is better.

\paragraph{Motion quality.} FID is the Fr\'echet distance between Gaussians fitted to the features of the
generated motions and of the real test motions.

\paragraph{Text alignment.} Evaluation texts and their generated motions are split into groups of 32.
R-precision R@$k$ is the fraction of texts whose own motion is among the $k$ motions of the group closest to the
text in feature space. MM-Dist is the mean Euclidean distance between a text's feature and the feature of the
motion generated for it. The CLIP score of \cref{tab:mardm} is the cosine similarity between a generated motion
and its caption under the CLIP-based evaluator of \citet{meng2025mardm}.

\paragraph{Diversity.} With $v$ the feature of a generated motion,
$\mathrm{Div}=\frac{1}{S_d}\sum_{i=1}^{S_d}\|v_i-v'_i\|_2$ over $S_d=300$ random pairs of generated motions, compared
with the value of real motions ($9.503$ on HumanML3D).
$\mathrm{MMod}=\frac{1}{CI}\sum_{c=1}^{C}\sum_{i=1}^{I}\|v_{c,i}-v'_{c,i}\|_2$ measures variation within a prompt:
$C=100$ prompts, 30 motions each, and $I=10$ random pairs per prompt.

\paragraph{Kimodo.} Texts and motions are embedded with Kimodo's released TMR
model~\citep{petrovich2023tmr}, trained for its skeleton. For each of the 607 prompts (BONES-SEED captions), R@$k$ ranks all 607 generated motions by similarity to
the prompt; a motion generated for a near-identical caption (TMR text-embedding cosine above $0.96$) also counts
as correct. FID is the Fr\'echet distance between the embeddings of the generated motions and of the captions in
TMR's joint space, so it uses no real motions and is not comparable with the HumanML3D FID. GT-R@1 is R@1
restricted to the 567 prompts that have a ground-truth motion.

\paragraph{Cost.} NFE counts guided denoising steps per motion, i.e., sampling steps: a classifier-free-guided
step counts once for every method, whether its conditional and unconditional branches run in one batched
call (HY-Motion, the latent-diffusion teachers MLD and MotionLCM-V2, and Kimodo) or in two (MDM, for teacher
and student alike). Time and memory are measured end to end, guidance included
(\cref{sec:latency}). The VLM-judged SSAE score is defined in \cref{app:ssae} and the human ranking in
\cref{app:human}.

\paragraph{Seeds and intervals.} Primary HumanML3D student rows average three training seeds ($42/43/44$).
On KIT-ML, \tacd\ point estimates also average three training seeds; the $\pm$ values beside evaluator
metrics are 95\% intervals over the 20 evaluator repetitions, not training-seed variation, and per-seed FID is
in \cref{tab:kit_steps} (eight-step standard deviation $0.008$). Gradient probes instead repeat minibatch
sampling on fixed weights.

\paragraph{Model selection.} Every trained row of \cref{tab:matched_lite,tab:kimodo}, ours and ported
alike, is the \emph{final iterate of a fixed iteration budget} (the artifact at \texttt{max\_iters}), not a
validation-FID-selected checkpoint.

\subsection{Training configurations}
\label{app:configs}
\Cref{tab:configs} names every training
configuration used in the paper.
\begin{table}[htbp]
\apptab
\centering
\begin{tabular}{@{}p{.18\linewidth}p{.19\linewidth}p{.06\linewidth}p{.33\linewidth}p{.175\linewidth}@{}}
\toprule
Purpose & Student / encoder & $K$ & Loss; sampling grid; margin $\delta$ & Replication \\
\midrule
Flow controls & 1B LoRA-32 / 8B & 4 & Uncapped/grid cap/\newline last-point;\newline uniform; $.001/.125$ & 3 each \\
\rowcolor{methodwash}
Lite controls & Lite full / 0.6B & 8 & Original/grid cap/\newline endpoint/last-point;\newline uniform; $.001/.0625$ & 3 each \\
Coefficient interventions & Lite full / 0.6B & 8 & Weight cap on original grid;\newline $C{=}10^6$ on capped grid & 3 each \\
Gradient probe & Lite full / 0.6B & 8 & Uncapped/grid cap;\newline uniform; $.001/.0625$ & 32 batches\newline $\times$ 2 probe seeds \\
\midrule
\rowcolor{methodwash}
DDIM transfer & Teacher-size / native & 4 & Endpoint matching;\newline native DDIM grid & MLD/V2: 3 \\
\bottomrule
\end{tabular}
\caption{\textbf{Experimental configurations.} HY-Motion 1B is the teacher for the first four rows;
replication counts denote training seeds except for gradient probes; text encoders remain frozen.
Teal rows contain the \tacd\ recipes: grid capping for the flow teacher, endpoint matching for
diffusion teachers.}
\label{tab:configs}
\end{table}

\label{sec:analysis}
\label{sec:limits}
\label{sec:limitations}
The terminal-amplification analysis is evaluated on the flow branch, where the endpoint-to-velocity
conversion induces the time-dependent coefficient; the diffusion experiments use the endpoint-matching
formulation of \cref{sec:ddim_instance}. The flow interventions are evaluated on uniform sampling grids at 1B/$K=4$ and Lite/$K=8$, with
three training seeds per configuration, and the bound is the closed-form midpoint rule
$\delta=1/(2K)$, $C=4K^2$.

\FloatBarrier %
\section{Baseline implementations}
\label{sec:ports}
We detail the ported baselines of \cref{tab:matched_lite}. Each port keeps its method-specific targets,
auxiliary networks, and hyperparameters, and shares the student architecture, prompt cache, seeds,
generator-update budget, and evaluation protocol.

The three port rows use the same student as the \tacd rows: the 0.46B Lite motion head with the
Qwen3-0.6B conditioning path and its initialization, the same 6,000-prompt cache, $K{=}8$, CFG $5.0$ at
sampling, AdamW with generator learning rate $3\times10^{-6}$, batch $2$ with $2$-step gradient
accumulation, 8,000 generator updates, seeds $42/43/44$, and the final checkpoint. Objective-specific settings are listed below.

\paragraph{DMD port.} \emph{Objective:} distribution matching~\citep{yin2024dmd,yin2024dmd2} ported to
the flow teacher. The generator's backward-simulated clean motion is trained against the difference
between the frozen teacher score and a learned fake score, with the fake network updated five times per
generator update at learning rate $10^{-5}$, together with the regression term of DMD on the teacher's
own (noise, 50-step output) pairs with weight $1$. No real motions enter either term.

\paragraph{Consistency port.} HY-Motion's training motions are not released, so the port first
synthesizes teacher motions and then applies consistency distillation~\citep{luo2023lcm} to that
corpus. It uses the teacher's 50-step CFG-$5$ outputs on the same prompt cache with fresh noise, $32$
consistency boundaries inside the eight-step grid, one CFG-$5$ teacher Euler substep per target, a
Consistency target EMA (decay $.95$), and a pseudo-Huber loss ($c=.01$). CFG is folded into the
student during training, so the student samples at CFG $1$, as in LCM.

\paragraph{Reflow port.} \emph{Objective:} rectified-flow distillation~\citep{liu2023rectified,liu2024instaflow}
on the teacher's own couplings. For each of the same 6,000 prompts, a noise sample $\epsilon$ and the
teacher's 50-step CFG-$5$ output $y_T$ define $x_t=(1-t)\epsilon+t\,y_T$, and the student's CFG-combined
velocity regresses $y_T-\epsilon$ at one stratified $t$ per segment of the eight-step grid. No real motions
enter. Every other setting is inherited from the \tacd\ rows unchanged (initialization, prompt cache,
learning rate, batch, 8,000 updates, EMA, seeds $42/43/44$, final checkpoint), so only the method-specific
training objective and target construction differ from \tacd.

\FloatBarrier %
\suppgroup{Results}
\section{Results by benchmark}
\label{sec:detailtabs}
This appendix gives the complete tables behind the quality comparisons of \cref{sec:overall,sec:extension}, one
subsection per benchmark: HumanML3D with the flow teacher (\cref{app:hml_flow}), HumanML3D with
diffusion teachers (\cref{app:hml_diff}), HumanML3D under a second evaluator (\cref{app:mardm}), and KIT-ML
(\cref{app:kit}); Kimodo follows in \cref{app:kimodo}.
\subsection{HumanML3D with the flow teacher}
\label{app:hml_flow}
Beyond \cref{tab:matched_lite}, this subsection gives published HumanML3D context and the teacher's
guidance sweep (\cref{tab:frontier}).

\paragraph{Published HumanML3D context.}
\Cref{tab:published_hml} retains the reported results of generators trained on real motions. MLCT and
MotionPCM are source-reported; the MotionLCM-V1 and V2 rows throughout this paper are the released
checkpoints generated by us and evaluated in our harness.

\begin{table}[htbp]
\apptab
\centering
\begin{tabular}{@{}llcccc@{}}
\toprule
Model & NFE & CFG & FID$\downarrow$ & R@3$\uparrow$ & MM-Dist$\downarrow$ \\
\midrule
Teacher (1B), 50 steps & 50 & 2.0 & 0.9937 & 0.7052 & 3.7308 \\
Teacher (1B), 50 steps & 50 & 3.0 & 0.6989 & 0.7312 & 3.5619 \\
Teacher (1B), 50 steps & 50 & 5.0 & 0.5407 & 0.7431 & 3.4706 \\
Teacher (1B), 50 steps & 50 & 7.5 & 0.5487 & 0.7408 & 3.4893 \\
\midrule
Teacher (1B), 8 steps (truncated) & 8 & 5.0 & 3.2503 & 0.6516 & 4.2250 \\
\midrule
\rowcolor{amplifiedwash}
Student (0.46B), uncapped (s42) & 8 & 5.0 & 2.5631 & 0.7056 & 3.7131 \\
\rowcolor{amplifiedwash}
Student (0.46B), uncapped (s43) & 8 & 5.0 & 2.4189 & 0.7082 & 3.6962 \\
\rowcolor{amplifiedwash}
Student (0.46B), uncapped (s44) & 8 & 5.0 & 2.4755 & 0.7089 & 3.6995 \\
\rowcolor{amplifiedwash}
\quad mean of the three seeds & 8 & 5.0 & 2.4859 & 0.7076 & 3.7029 \\
\rowcolor{methodwash}
TACD, 0.46B (s42) & 8 & 5.0 & 1.0085 & 0.7829 & 3.1811 \\
\rowcolor{methodwash}
TACD, 0.46B (s43) & 8 & 5.0 & 1.0737 & 0.7813 & 3.1999 \\
\rowcolor{methodwash}
TACD, 0.46B (s44) & 8 & 5.0 & 1.0399 & 0.7821 & 3.1955 \\
\rowcolor{methodwash}
\quad mean of the three seeds & 8 & 5.0 & 1.0407 & 0.7821 & 3.1922 \\
Endpoint matching, 0.46B (3-seed mean) & 8 & 5.0 & 1.0292 & 0.7812 & 3.1876 \\
\midrule
Real motions (reference) & -- & -- & -- & 0.8075 & 2.9453 \\
\bottomrule
\end{tabular}

\caption{\textbf{Retrieval--quality trade-off: eight-step Lite vs.\ the 50-step HY-Motion teacher.}
Student rows: Qwen3-0.6B encoder, full 0.46B Lite backbone; teacher rows: its 8B encoder. Mean rows
average three seeds; CFG $5.0$ is the teacher's best FID and R@3 of four tested. Rose rows: terminal
coefficient $10^6$.}
\label{tab:frontier}
\end{table}

\paragraph{Teacher guidance sweep.}
\tacd's three-seed R@3 is $0.7821$ (range $0.7813$--$0.7829$), MM-Dist is
$3.1922$ ($3.1811$--$3.1999$), FID is $1.0407$ ($1.0085$--$1.0737$), and diversity is
$8.9186$ ($8.8859$--$8.9444$). The corresponding teacher readings are
$0.7431$, $3.4706$, $0.5407$, and $9.1315$. The teacher's best observed FID and R@3 both use
CFG $5.0$ over $\{2.0,3.0,5.0,7.5\}$; \cref{tab:frontier} preserves the complete sweep.

\subsection{Diffusion teachers on HumanML3D}
\label{app:hml_diff}
\Cref{tab:context} puts both diffusion teachers next to their released students on one metric
protocol, and \cref{tab:mld} gives the full MLD and MotionLCM-V2-teacher comparisons. At matched NFE ($K{=}4$) \tacd's FID is
below the V2 student's ($0.055$ vs.\ $0.072$) with higher MultiModality.
\begin{table}[htbp]
\apptab
\begin{minipage}[t]{0.46\linewidth}
\vspace{0pt}
\centering
\emph{(a) Original MLD teacher (the teacher MotionLCM-V1 distills)}\\[2pt]
\resizebox{\linewidth}{!}{%
\begin{tabular}{@{}lcccc@{}}
\toprule
Method & NFE & FID$\downarrow$ & R@3$\uparrow$ & Div$\rightarrow$ \\
\midrule
Teacher (MLD, 50)            & 50 & 0.387\makebox[18pt]{} & 0.795 & 9.53 \\
Truncated teacher            & 4  & 0.585\makebox[18pt]{} & 0.768 & 9.49 \\
\midrule
MotionLCM-V1 (consistency)   & 4  & 0.286\makebox[18pt]{} & 0.798 & --   \\
\rowcolor{methodwash}
\tacd                        & 4  & 0.222\sd{.002} & 0.818 & 9.80 \\
\bottomrule
\end{tabular}}
\end{minipage}\hfill
\begin{minipage}[t]{0.53\linewidth}
\vspace{0pt}
\centering
\emph{(b) Retrained V2 teacher (the teacher of the V2 student)}\\[2pt]
\resizebox{\linewidth}{!}{%
\begin{tabular}{@{}lccccc@{}}
\toprule
Method & NFE & FID$\downarrow$ & R@3$\uparrow$ & Div$\rightarrow$ & MMod$\uparrow$ \\
\midrule
MotionLCM-V2 (consistency) & 4 & 0.072\makebox[18pt]{} & 0.841 & 9.54 & 1.72 \\
\midrule
\rowcolor{methodwash}
\tacd         & 4 & 0.055\sd{.001} & 0.824 & 9.57 & 1.94 \\
\bottomrule
\end{tabular}}
\end{minipage}
\caption{\textbf{MotionLCM comparison}: \emph{(a)} the MLD teacher, \emph{(b)} the MotionLCM-V2 teacher.
MotionLCM rows: released checkpoints, our harness. \tacd\ rows: three-seed mean$\pm$std.}
\label{tab:mld}
\label{tab:v2}
\end{table}

\subsection{A second HumanML3D evaluator}
\label{app:mardm}
\Cref{tab:mardm} rescores the generated motions of \cref{tab:matched_lite,tab:context} with the released
67-dimensional evaluators and evaluation code of \citet{meng2025mardm} over 20 repetitions, keeping each test
item's caption and length from our Guo protocol; on real test motions the loader reproduces the reference values
of \citet{meng2025acmdm} (R@3 $.795$; MM-Dist $3.241$ vs.\ $3.244$). Under this evaluator \tacd\ (HY-Motion)
remains the best eight-step system on every metric in each of its three seeds.

\subsection{KIT-ML}
\label{app:kit}
\begin{table}[htbp]
\scriptsize
\renewcommand{\arraystretch}{1.0}
\setlength{\tabcolsep}{3pt}
\begin{minipage}[t]{0.545\linewidth}
\vspace{0pt}
\centering
\resizebox{\linewidth}{!}{%
\begin{tabular}{@{}lcccccc@{}}
\toprule
Method & NFE & R@1$\uparrow$ & R@3$\uparrow$ & MM-Dist$\downarrow$ & FID$\downarrow$ & Div$\rightarrow$ \\
\midrule
\multicolumn{7}{@{}l}{\emph{(a) Published generators (real-motion training)}} \\
Real motions~\citep{guo2022humanml3d}        & --         & .424   & .779  & 2.788  & .031  & 11.08 \\
T2M~\citep{guo2022humanml3d}                 & --         & .361   & .681  & 3.488  & 3.022 & 10.72 \\
MotionDiffuse~\citep{zhang2023motiondiffuse} & --         & .417   & .739  & 2.958  & 1.954 & 11.10 \\
MDM$^{\dagger}$~\citep{tevet2023mdm}         & $\sim$1000 & .404   & .731  & 3.096  & .513  & 10.73 \\
MLD~\citep{chen2023mld}                      & 50         & .390   & .734  & 3.204  & .404  & 10.80 \\
T2M-GPT~\citep{guo2022t2mgpt}                & --         & .416   & .745  & 3.007  & .514  & 10.92 \\
ReMoDiffuse~\citep{zhang2023remodiffuse}     & --         & .427   & .765  & 2.814  & .155  & 10.80 \\
MotionGPT~\citep{jiang2023motiongpt}         & --         & .366   & .680  & 3.527  & .510  & 10.35 \\
MoMask~\citep{guo2024momask}                 & --         & .433   & .781  & 2.779  & .204  & --    \\
MotionPCM~\citep{motionpcm2025}              & 4          & .443   & .789  & 2.881  & .336  & 10.76 \\
IRG-MotionLLM S3~\citep{irg_motionllm}       & --         & .445   & .781  & 2.740  & .432  & 11.12 \\
\bottomrule
\end{tabular}}
\end{minipage}\hfill
\begin{minipage}[t]{0.435\linewidth}
\vspace{0pt}
\centering
\resizebox{\linewidth}{!}{%
\begin{tabular}{@{}lrrrrrr@{}}
\toprule
Method & NFE & R@1$\uparrow$ & R@3$\uparrow$ & MM-Dist$\downarrow$ & FID$\downarrow$ & Div$\rightarrow$ \\
\midrule
\multicolumn{7}{@{}l}{\emph{(b) Our evaluation: MDM teacher, teacher-only students}} \\
Real motions (GT)          & --   & .426 & .780 & 2.786 & .027  & 10.99 \\
MDM teacher                & 1000 & .403 & .725 & 3.115 & .506  & 10.74 \\
\addlinespace[6.6pt]
Truncated teacher          & 2    & .395 & .728 & 3.193 & 1.530 & 11.24 \\
\rowcolor{methodwash}
\tacd                      & 2    & .410 & .756 & 2.996 & 1.038 & 11.06 \\
\addlinespace[6.6pt]
Truncated teacher          & 4    & .410 & .755 & 3.014 & .976  & 11.38 \\
\rowcolor{methodwash}
\tacd                      & 4    & .416 & .761 & 2.932 & .852  & 11.16 \\
\addlinespace[6.6pt]
Truncated teacher          & 8    & .416 & .756 & 2.948 & .745  & 11.34 \\
\rowcolor{methodwash}
\tacd                      & 8    & .421 & .763 & 2.873 & .649  & 11.08 \\
\bottomrule
\end{tabular}}
\end{minipage}
\caption{\textbf{KIT-ML, MDM teacher.} \tacd\ beats the truncated teacher at every NFE in R@1, R@3, MM-Dist, and FID; \tacd\ rows: three-seed means after 10k updates; intervals in \cref{tab:kit_steps}. $^{\dagger}$Post-fix evaluation endorsed by the MDM authors (repository issue \#211).}
\label{tab:kit}
\end{table}

\Cref{tab:kit}b evaluates the released MDM teacher~\citep{tevet2023mdm}, an $x_0$-prediction,
251-d raw-motion diffusion model, with TACD's DDIM endpoint-matching recipe at two, four, and eight
steps. The measured block uses one evaluation harness: the 1000-step teacher has
FID $0.506\pm0.023$ and R@3 $0.725\pm0.005$. Few-step samplers use trailing timestep
spacing~\citep{lin2024common}, which places the first step at $t=999$. At the same NFE, endpoint
matching reduces FID by $32\%$ at two steps and $13\%$ at four and eight.

\paragraph{Evaluation conventions.}
Published rows (\cref{tab:kit}a) retain their sources' conventions. Our evaluation takes the caption length
before padding, the corrected path of the released MDM code; the re-measured teacher row is within about one
reported confidence interval of the corrected MDM reference on all four checked metrics.
\Cref{tab:kit_steps} repeats the measured rows with their 95\% intervals over the 20 evaluator repetitions and
the per-seed FID of the \tacd\ rows. GT is recomputed in each run; the displayed GT row comes from the teacher run.
\begin{table}[htbp]
\apptab
\centering
\begin{tabular}{@{}lrrrrrrl@{}}
\toprule
Method & NFE & R@1$\uparrow$ & R@3$\uparrow$ & MM-Dist$\downarrow$ & FID$\downarrow$ & Div$\rightarrow$ & FID by seed \\
\midrule
Real motions (GT)   & --   & .426 & .780\sd{.005} & 2.786 & .027\sd{.004}  & 10.99 & -- \\
MDM teacher         & 1000 & .403 & .725\sd{.005} & 3.115 & .506\sd{.023}  & 10.74 & -- \\
\addlinespace[2pt]
Truncated teacher   & 2    & .395 & .728\sd{.005} & 3.193 & 1.530\sd{.063} & 11.24 & -- \\
\rowcolor{methodwash}
\tacd               & 2    & .410 & .756\makebox[18pt]{} & 2.996 & 1.038\makebox[18pt]{} & 11.06 & 1.020 / 1.077 / 1.017 \\
\addlinespace[2pt]
Truncated teacher   & 4    & .410 & .755\sd{.006} & 3.014 & .976\sd{.064}  & 11.38 & -- \\
\rowcolor{methodwash}
\tacd               & 4    & .416 & .761\makebox[18pt]{} & 2.932 & .852\makebox[18pt]{}  & 11.16 & .840 / .857 / .859 \\
\addlinespace[2pt]
Truncated teacher   & 8    & .416 & .756\sd{.006} & 2.948 & .745\sd{.051}  & 11.34 & -- \\
\rowcolor{methodwash}
\tacd               & 8    & .421 & .763\makebox[18pt]{} & 2.873 & .649\makebox[18pt]{}  & 11.08 & .649 / .656 / .641 \\
\bottomrule
\end{tabular}
\caption{\textbf{KIT-ML measured rows with evaluator intervals} (the rows of \cref{tab:kit}; Guo evaluator,
20 repetitions). $\pm$: 95\% interval over the 20 repetitions. \tacd\ rows average seeds $42/43/44$; the last column lists their per-seed FID.}
\label{tab:kit_steps}
\end{table}

\section{Kimodo}
\label{app:kimodo}
The main-text Kimodo result is in \cref{sec:kimodo_main,tab:kimodo_main}; this appendix gives the
full track, the encoder comparison, and a held-out split analysis. Kimodo is a 100-step
$x_0$-diffusion teacher over a 77-joint skeleton, trained on BONES-SEED; \tacd\ uses none of its motions.
All comparisons use a 607-prompt BONES-SEED caption set scored with Kimodo's released TMR evaluator, and GT-R@1 is R@1 restricted to the 567 prompts that have a ground-truth
motion.
\subsection{Step compression and conditioning}
\label{sec:kimodo}
\Cref{tab:kimodo} gives the full Kimodo track: \tacd\ with the teacher's 8B encoder at $K\in\{8,4,2\}$,
and the teacher's own DDIM at the same budgets; \cref{tab:kimodo_main} adds \tacd\ (Kimodo).

\begin{table}[htbp]
\apptab
\centering
\begin{tabular}{@{}lrrrrrrr@{}}
\toprule
Method & NFE & \multicolumn{1}{c}{R@1 (\%)$\uparrow$} & \multicolumn{1}{c}{R@3 (\%)$\uparrow$} & FID$\downarrow$ & \multicolumn{1}{c}{GT-R@1 (\%)$\uparrow$} & Lat.(s)$\downarrow$ & VRAM \\
\midrule
Teacher                & 100 & 73.15\makebox[18pt]{} & 88.63\makebox[18pt]{} & 0.045 & 75.66\makebox[18pt]{} & 2.96 & 15.3G \\
Truncated teacher      & 8   & 69.36\makebox[18pt]{} & 84.35\makebox[18pt]{} & 0.064 & 72.66\makebox[18pt]{} & 0.29 & 15.3G \\
Truncated teacher      & 4   & 67.05\makebox[18pt]{} & 82.37\makebox[18pt]{} & 0.079 & 70.55\makebox[18pt]{} & 0.18 & 15.3G \\
Truncated teacher      & 2   & 60.46\makebox[18pt]{} & 77.27\makebox[18pt]{} & 0.107 & 63.14\makebox[18pt]{} & 0.13 & 15.3G \\
\rowcolor{methodwash}
\tacd, 8B enc.         & 8   & 71.39\sd{.50} & 86.55\sd{.25} & 0.054 & 75.01\sd{.37} & 0.29 & 15.3G \\
\rowcolor{methodwash}
\tacd, 8B enc.         & 4   & 68.42\sd{.69} & 83.25\sd{.38} & 0.067 & 72.13\sd{.77} & 0.18 & 15.3G \\
\rowcolor{methodwash}
\tacd, 8B enc.         & 2   & 62.36\sd{.12} & 79.57\sd{.23} & 0.095 & 65.43\sd{.00} & 0.13 & 15.3G \\
\bottomrule
\end{tabular}
\caption{\textbf{Kimodo results} (607-caption TMR evaluation; GT-R@1: R@1 restricted to the 567 prompts that
have a ground-truth motion). $\pm$: SD over three seeds ($K{=}2$: two). \tacd\ rows: teacher's 8B encoder. Lat./VRAM: end to end,
one RTX L40.}
\label{tab:kimodo}
\end{table}

\begin{table}[t]
\scriptsize
\setlength{\tabcolsep}{3pt}
\renewcommand{\arraystretch}{1.0}
\begin{minipage}[c]{0.59\linewidth}
\centering
\resizebox{\linewidth}{!}{%
\begin{tabular}{@{}llrrrrrr@{}}
\toprule
Method & Encoder & NFE & R@1 (\%)$\uparrow$ & R@3 (\%)$\uparrow$ & FID$\downarrow$ & Time (s)$\downarrow$ & Memory (GB)$\downarrow$ \\
\midrule
Teacher & 8B & 100 & 73.15 & 88.63 & 0.045 & 2.96 & 15.3 \\
\addlinespace[2pt]
Truncated teacher & 8B & 8 & 69.36 & 84.35 & 0.064 & 0.29 & 15.3 \\
\rowcolor{methodwash}
\tacd & 8B & 8 & 71.39 & 86.55 & 0.054 & 0.29 & 15.3 \\
\rowcolor{methodwash}
\tacd\ (Kimodo) & 1B & 8 & 68.86 & 86.66 & 0.054 & 0.25 & 4.0 \\
\addlinespace[2pt]
Truncated teacher & 8B & 4 & 67.05 & 82.37 & 0.079 & 0.18 & 15.3 \\
\rowcolor{methodwash}
\tacd & 8B & 4 & 68.42 & 83.25 & 0.067 & 0.18 & 15.3 \\
\rowcolor{methodwash}
\tacd\ (Kimodo) & 1B & 4 & 67.00 & 82.98 & 0.066 & 0.13 & 4.0 \\
\addlinespace[2pt]
Truncated teacher & 8B & 2 & 60.46 & 77.27 & 0.107 & 0.13 & 15.3 \\
\rowcolor{methodwash}
\tacd & 8B & 2 & 62.36 & 79.57 & 0.095 & 0.13 & 15.3 \\
\bottomrule
\end{tabular}}
\end{minipage}\hfill
\begin{minipage}[c]{\dimexpr0.97\linewidth-0.59\linewidth\relax}
\caption{\textbf{Kimodo teacher} (607-caption TMR evaluation). \tacd\ with the 8B encoder beats the truncated teacher at every NFE; at NFE 8, \tacd\ (Kimodo) with a 1B text encoder keeps R@3 and FID with $3.8\times$ less memory (paired runs in \cref{tab:item7}).}
\label{tab:kimodo_main}
\label{tab:efficiency}
\end{minipage}
\end{table}

Step compression transfers: \tacd at
$K{=}8$ reaches R@1 $71.39{\pm}0.50$ / R@3 $86.55{\pm}0.25$ / FID $0.0542$ against the teacher's $73.15$ /
$88.63$ / $0.045$. The
teacher's own DDIM degrades gracefully---$69.36$ / $67.05$ / $60.46$ at $K{=}8$ / $4$ / $2$---and
\tacd leads it by $+2.03$ / $+1.37$ / $+1.89$ R@1, and each of its eight training runs (three seeds at
$K=8/4$, two at $K=2$) exceeds the truncated teacher at the same step count in R@1, R@3, and FID.

The conditioning axis transfers too: a student conditioned on Llama-3.2-1B through a distilled
layer-fusion head (no motion data anywhere) cuts VRAM $15.3{\to}4.0$\,GB. The two efficiency axes
separate: of the end-to-end $11.9\times$ against the 100-step teacher ($2.96$\,s$\to$$0.25$\,s), step
compression accounts for $10.1\times$ ($2.96$\,s$\to$$292$\,ms with the 8B encoder at the student's own
step count) and the encoder change for $1.18\times$ ($292{\to}248$\,ms); at $K{=}8$ the denoiser, not
the text encoder, is most of the wall clock. VRAM is entirely the encoder ($3.8\times$), since step
count does not move a sampler's peak. All timings are medians over five round-robin passes on one RTX
L40 (\cref{sec:latency}). \Cref{tab:item7}
trains the student on the 1B encoder outright, against the 8B encoder under matched seeds.

\subsection{Training on the 1B encoder}
\label{app:kimodo_small}
The \tacd\ rows of
\cref{tab:kimodo} were trained with the teacher's own 8B encoder. \Cref{tab:item7} trains the student
end-to-end on the encoder it runs with: three seeds per cell at $K{=}8$, with the same code,
schedule, and generation settings.

\begin{table}[htbp]
\apptab
\centering
\renewcommand{\rng}[1]{\textcolor{tableink}{[#1]}}%
\resizebox{\linewidth}{!}{%
\begin{tabular}{@{}lccccc@{}}
\toprule
Conditioning encoder & NFE & \multicolumn{1}{c}{R@1 (\%)$\uparrow$} & \multicolumn{1}{c}{R@3 (\%)$\uparrow$} & \multicolumn{1}{c}{FID$\downarrow$} & \multicolumn{1}{c}{GT-R@1 (\%)$\uparrow$} \\
\midrule
teacher's own 8B encoder & 8 & 71.72\sd{.50} & 86.66\sd{.33} & 0.0542\sd{.0002} & 75.25\sd{.10} \\
\rowcolor{methodwash}
1B encoder, trained in & 8 & 68.86\sd{.44} & 86.66\sd{.28} & 0.0544\sd{.0003} & 72.60\sd{.67} \\
\quad paired $\Delta$ (1B $-$ 8B) & 8 & $-2.86$ \rng{$-3.30$, $-2.47$} & $-0.00$ \rng{$-0.66$, $+0.49$} & $+0.0002$ \rng{$-0.0001$, $+0.0004$} & $-2.65$ \rng{$-3.18$, $-1.94$} \\
\bottomrule
\end{tabular}}
\caption{\textbf{Training on the 1B text encoder vs.\ the teacher's own 8B encoder} (Kimodo, 607-prompt
bench, $K{=}8$; the two conditions share seeds $42/43/44$). $\pm$: SD over
three seeds; brackets on the $\Delta$ rows: range of the three paired differences.}
\label{tab:item7}
\end{table}

Swapping the 8B encoder for the
trained-in 1B encoder leaves FID unchanged ($+0.0002$, against per-seed standard deviations of at most
$0.0003$) and R@3 flat. The 8B-encoder rows of \cref{tab:item7} are separate training runs
from those of \cref{tab:kimodo}.

\subsection{Held-out split analysis}
\label{app:kimodo_overlap}
Of the 567 GT-bearing prompts, 462 caption BONES-SEED clips in the teacher's training split and 105 are
held out; these clips are used only for evaluation, never for student training. Restricted to each stratum,
\tacd@8 and the truncated teacher@8 score $74.0$/$71.9$ (teacher-train) and $88.6$/$84.8$ (held-out) GT-R@1:
on the held-out prompts \tacd\ keeps its advantage ($+3.8$\,pp, against $+2.2$\,pp on the rest).

\section{Semantic fidelity under a VLM judge}
\label{app:ssae}
This appendix details the VLM-judged comparison of \cref{sec:overall}, which scores semantic fidelity
through questions about the described motion rather than through a learned embedding. It follows the SSAE
construction of HY-Motion~\citep{hymotion2025}, with our own prompt set and judge models.
\Cref{tab:ssae} scores eight HY-Motion systems and ten open-weight HumanML3D models with a
vision--language judge on a fixed set of 640 prompts in six categories: 600 captions of BONES-SEED motion clips,
tagged generic, compositional, common, long, or rare-verb, and 40 prompts we wrote outside that corpus. None
comes from HumanML3D, whose texts the HY-Motion students train on. The released Lite row uses its own Qwen3-8B encoder
at 50 steps; the ports are the students of \cref{tab:matched_lite}, the
Consistency port sampled at its own CFG of 1.

\paragraph{Open-weight models.} Each model uses its released HumanML3D checkpoint with its own default
sampler and guidance, one fixed seed, and six-second clips: MDM~\citep{tevet2023mdm} at 1000 steps, the MLD
teacher of \cref{tab:context}~\citep{chen2023mld} with 50 DDIM steps, T2M-GPT~\citep{guo2022t2mgpt},
MoMask~\citep{guo2024momask}, MARDM~\citep{meng2025mardm}, MotionHiFlow~\citep{motionhiflow},
MoGeFlow-L~\citep{moge_flow} with 96 flow steps, and the four-step MotionLCM~\citep{dai2024motionlcm},
MLCT~\citep{hu2024mlct}, and MotionLCM-V2~\citep{dai2024motionlcmv2}. Their HumanML3D features are decoded
to joint positions, retargeted onto the HY-Motion skeleton by matching bone directions (mean error
below $4^\circ$), and rendered with the same renderer, camera, and resolution as the HY-Motion rows;
the last two frames of MARDM outputs, which contain a decoding jump, are held at the preceding frame. For this
block \tacd\ goes through the same path ($.699$; $.720$ when rendered directly), so the lower block of
\cref{tab:ssae} compares against \tacd\ rendered from joints.

\paragraph{Protocol.} For each prompt, gpt-5.5 authored binary sub-questions about the described
motion (``does the person raise both arms?''), and gpt-5.4-mini gated them by answerability from
video. Each generation was rendered as a six-second clip, of which the judge (gpt-5.6-luna) saw 14
frames spread over the clip (384\,px); it answered every question three times, a question counting when all three
votes are usable and its answer being the majority vote. A system's score is its yes-rate over the
2,571 gated questions (636 prompts) that all eight HY-Motion systems cover; differences are paired by prompt with
a 4,000-sample prompt-cluster bootstrap.

\FloatBarrier %
\section{Human ranking study}
\label{app:human}
The study behind \cref{tab:human} compares four systems on 50 prompts drawn at random from the 640 of \cref{app:ssae},
stratified by prompt category: eight-step \tacd, the 50-step HY-Motion teacher, and the four-step
open-weight models MotionLCM-V2 and MLCT, all rendered from joints as in the lower block of
\cref{tab:ssae}. Each page shows the prompt and the four
six-second clips playing together; the systems are hidden and their screen positions are balanced
within each annotator. Annotators rank the clips primarily by how well the motion follows the prompt,
using motion naturalness only to break ties. \Cref{tab:human} reports, for each
system, the share of rankings that place it above the teacher, and the prompts on which most annotators
place it above (won), below (lost), or equally often (tied). 17 annotators contributed 691 rankings;
confidence intervals come from 4,000 bootstrap samples that resample annotators and prompts together.

\FloatBarrier %
\section{Controlled interventions and gradient statistics}
\label{app:diagrows}
This appendix gives the per-seed records and gradient statistics behind \cref{sec:onestate}: the
supervision-grid control on the 1B student (\cref{app:gridcontrol}), the coefficient interventions by
seed (\cref{app:coefseeds}), the gradient probes and the role of clipping (\cref{app:gradstats}), and
the full terminal-coefficient sweep (\cref{app:coefsweep}).
\subsection{Supervision-grid control on the 1B student}
\label{app:gridcontrol}
\Cref{tab:ablation} repeats the supervision-grid control of \cref{sec:repair} on the 1B student at
$K{=}4$: the seed-matched uncapped and capped pair (three seeds each, identical except for $\delta$) and a
last-point-only control, all trained and evaluated under the main protocol. Capping the grid lowers mean FID from $3.291$ to $1.458$.

\begin{table}[htbp]
\apptab
\ifdefined\tacdpaira\else\newsavebox{\tacdpaira}\fi\sbox{\tacdpaira}{\begin{tabular}{@{}lr@{}}
\toprule
Supervision grid & FID$\downarrow$ \\
\midrule
\rowcolor{amplifiedwash}
Original (uncapped) & 3.291 \\
\rowcolor{methodwash}
Capped (\tacd) & 1.458 \\
Last point only & 1.462 \\
\bottomrule
\end{tabular}}
\ifdefined\tacdpairb\else\newsavebox{\tacdpairb}\fi\sbox{\tacdpairb}{\begin{tabular}{@{}lrrr@{}}
\toprule
Training configuration & Seed 42 & Seed 43 & Seed 44 \\
\midrule
Original, $C=256$ (weight capping) & 1.0437 & 1.0098 & 0.9798 \\
\rowcolor{amplifiedwash}
Capped, $C=10^6$ & 2.6593 & 2.6454 & 2.7029 \\
\bottomrule
\end{tabular}}
\edef\tacdwa{\the\wd\tacdpaira}\edef\tacdwb{\the\wd\tacdpairb}\edef\tacdlw{\the\linewidth}\pgfmathsetmacro{\tacdpairscale}{min(1, 0.42*\tacdlw/\tacdwa, 0.54*\tacdlw/\tacdwb)}
\begin{minipage}[c]{0.42\linewidth}
\centering
\scalebox{\tacdpairscale}{\usebox{\tacdpaira}}
\end{minipage}\hfill
\begin{minipage}[c]{0.54\linewidth}
\centering
\scalebox{\tacdpairscale}{\usebox{\tacdpairb}}
\end{minipage}
\par
\begin{minipage}[t]{0.42\linewidth}
\caption{\rightskip=0pt plus 2em\relax \textbf{HY-Motion 1B/$K{=}4$ supervision-grid control.} Three-seed mean FID; uniform segment
weights. Rose row: terminal coefficient $10^6$.}
\label{tab:ablation}
\end{minipage}\hfill
\begin{minipage}[t]{0.54\linewidth}
\caption{\rightskip=0pt plus 2em\relax \textbf{Direct coefficient interventions by training seed.}
Entries are FID; the first row is weight capping (\circnum{3}) of \cref{tab:controls}, the second
restores the terminal coefficient on the capped grid.}
\label{tab:coefficient_seeds}
\end{minipage}
\end{table}

\emph{Capped} ($\delta{=}1/(2K)$, $w_k{\equiv}1$): FID $1.418/1.417/1.540$, mean
$1.458$ (range $1.417$--$1.540$); R@3 $0.735/0.734/0.732$; Div $8.25/8.26/8.21$. \emph{Uncapped} ($\delta{=}10^{-3}$,
$w_k{\equiv}1$): FID $3.233/3.305/3.335$, mean $3.291$ (range $3.233$--$3.335$);
R@3 $0.689/0.684/0.686$; Div $7.73/7.72/7.70$.
With $M{=}2$ supervision times per segment the cap moves both of the final segment's supervision
times ($0.8745,0.999 \to 0.8125,0.875$). The last-point-only control separates the two: with the grid built on the
original knots and only the final supervision time clamped ($0.8745,0.875$), FID is $1.462$ (3 seeds,
$1.418/1.425/1.544$, range $1.418$--$1.544$), so moving the last supervision time alone is sufficient
in this configuration; it matches the capped arm seed for seed ($-0.001/{+}0.008/{+}0.004$).

\subsection{Coefficient interventions by seed}
\label{app:coefseeds}
\Cref{tab:coefficient_seeds} gives, seed by seed, the two single-coefficient interventions of
\cref{sec:repair}: lowering the terminal coefficient on the original grid, and restoring it on the
capped grid.

Weight capping (\circnum{3} in \cref{tab:controls}) is the first row of \cref{tab:coefficient_seeds};
its three-seed mean FID is $1.011$. Restoring $10^6$ on the capped grid gives $2.6593$, $2.6454$, and $2.7029$ at seeds 42, 43, and 44
(mean $2.669$).

\subsection{Gradient probes and clipping}
\label{app:gradstats}
The gradient shares and norms of \cref{sec:onestate} come from the probes and training logs described
here.
\paragraph{Full-backbone gradient probes.}
\label{app:gradient_probe}
We compute the terminal segment's gradient norm divided by the sum of all segment gradient norms
within each minibatch, then average those shares over 32 minibatches (a mean of ratios).
At initialization and $K=8$, the terminal segment carries $68.3\%$ of the gradient on the original
supervision grid and $0.86/0.89\%$ on the capped grid for probe seeds $42/43$. The eight final-checkpoint cells in \cref{tab:gradient_cross} use those same
probe seeds and 32 minibatches each, loading \texttt{full\_ema} from the 8,000-iteration
training-seed-42 checkpoint of each arm. All eight use the full 465.041M backbone.
\begin{table}[htbp]
\apptab
\ifdefined\tacdpaira\else\newsavebox{\tacdpaira}\fi\sbox{\tacdpaira}{\begin{tabular}{@{}lcc@{}}
\toprule
\multicolumn{3}{@{}l}{\emph{Final-segment share of gradient norm (\%), seeds 42\,/\,43}} \\
Grid probed & Uncapped ckpt. & \tacd\ ckpt. \\
\midrule
Original & 96.62\,/\,96.64 & 97.60\,/\,97.10 \\
Capped & 10.53\,/\,11.77 & 4.28\,/\,3.64 \\
\midrule
\multicolumn{3}{@{}l}{\emph{Pre-clip gradient norm, training seeds 42\,/\,43\,/\,44}} \\
Configuration & Median norm & Clipped (of 800) \\
\midrule
Uncapped student & 245.0\,/\,251.8\,/\,263.4 & all \\
\tacd\ (grid capping) & 5.97\,/\,6.06\,/\,6.03 & all \\
Endpoint matching & 5.12\,/\,5.04\,/\,4.96 & 799\,/\,800\,/\,800 \\
Weight capping & 6.05\,/\,5.91\,/\,6.06 & all \\
Last-point-only & 5.92\,/\,6.10\,/\,6.05 & all \\
Capped grid, $C=10^6$ & 527.4\,/\,533.6\,/\,541.3 & all \\
\bottomrule
\end{tabular}}
\ifdefined\tacdpairb\else\newsavebox{\tacdpairb}\fi\sbox{\tacdpairb}{\setlength{\tabcolsep}{2pt}\begin{tabular}{@{}llrrrrrr@{}}
\toprule
Family & Coefficient & FID$\downarrow$ & R@1$\uparrow$ & R@3$\uparrow$ & MM-Dist$\downarrow$ & Div$\rightarrow$ & Norm \\
\midrule
\multicolumn{8}{@{}l}{\emph{Weight family: original grid (queries $.937,\ .999$), terminal coefficient set to $C$ by \cref{eq:intervention}}} \\
Weight & $C=256$ ($4K^2$)   & 1.0111 & .4800 & .7814 & 3.190 & 8.925 & 6.01 \\
Weight & $C=1024$           & 1.0379 & .4792 & .7823 & 3.180 & 8.889 & 5.88 \\
Weight & $C=4096$           & 1.0671 & .4802 & .7823 & 3.172 & 8.879 & 6.24 \\
Weight & $C=16384$          & 1.1356 & .4754 & .7797 & 3.188 & 8.820 & 7.77 \\
Weight & $C=65536$          & 1.2995 & .4707 & .7769 & 3.221 & 8.640 & 17.82 \\
Weight & $C=262144$         & 1.6985 & .4457 & .7559 & 3.367 & 8.384 & 67.59 \\
\rowcolor{amplifiedwash}
Weight & $C=10^6$ (uncapped) & 2.4859 & .4024 & .7076 & 3.703 & 8.005 & 253.4 \\
\midrule
\multicolumn{8}{@{}l}{\emph{Grid family: last query moved to $1-\delta$, coefficient $\delta^{-2}$}} \\
Grid & $\delta=1/32$ ($b=1024$)  & 0.9955 & .4834 & .7856 & 3.172 & 8.932 & 5.91 \\
Grid & $\delta=3/32$ ($b=113.8$) & 1.0329 & .4794 & .7828 & 3.177 & 8.914 & 6.01 \\
\rowcolor{methodwash}
Grid & $\delta=1/16$ ($b=256$, TACD) & 1.041$^{\ast}$ & -- & -- & -- & -- & 6.26 \\
\bottomrule
\end{tabular}}
\edef\tacdwa{\the\wd\tacdpaira}\edef\tacdwb{\the\wd\tacdpairb}\edef\tacdlw{\the\linewidth}\pgfmathsetmacro{\tacdpairscale}{min(1, 0.48*\tacdlw/\tacdwa, 0.49*\tacdlw/\tacdwb)}
\begin{minipage}[c]{0.48\linewidth}
\centering
\scalebox{\tacdpairscale}{\usebox{\tacdpaira}}
\end{minipage}\hfill
\begin{minipage}[c]{0.49\linewidth}
\centering
\scalebox{\tacdpairscale}{\usebox{\tacdpairb}}
\end{minipage}
\par
\begin{minipage}[t]{0.48\linewidth}
\caption{\rightskip=0pt plus 2em\relax \textbf{Gradient statistics on the eight-step Lite student.} Top: final-segment share of the summed
segment-gradient norms, by probe grid and checkpoint. Bottom: median pre-clip gradient norm
and the number of the 800 logged updates on which the unit-norm clip binds.}
\label{tab:gradient_cross}
\label{tab:gnorm}
\end{minipage}\hfill
\begin{minipage}[t]{0.49\linewidth}
\caption{\rightskip=0pt plus 2em\relax \textbf{Terminal-coefficient sweep on Lite/$K{=}8$.} Each arm is the recipe of
\cref{tab:controls} with one flag changed (Guo evaluator, 20 repetitions). Norm: median pre-clip
gradient norm over 800 logged updates. $^{\ast}$Three-seed mean of \cref{tab:frontier} (range $1.009$--$1.074$).}
\label{tab:coefsweep}
\end{minipage}
\end{table}

\paragraph{Gradient magnitude is normalized by clipping.}
\label{app:gnorm}
All Lite/$K=8$ arms of \cref{tab:controls} train with AdamW and gradient clipping at unit norm.
\Cref{tab:gnorm} reports the pre-clip gradient norm, logged every ten updates. At the logged updates the
clip is active in essentially every observation (all 800 in 17 of the 18 runs, 799 in the remaining one),
so the compared configurations share the same clipping regime; the arms differ in how that gradient is composed across
segments, which \cref{tab:gradient_cross} measures as the terminal share.

\subsection{Terminal-coefficient sweep}
\label{app:coefsweep}
\Cref{tab:coefsweep} lists the arms behind \cref{fig:coefsweep}.

Each arm is the recipe of
\cref{tab:controls} with one flag changed: the weight family keeps the original grid (queries $.937$
and $.999$) and sets the terminal coefficient to $C$ by \cref{eq:intervention}; the grid family moves
the last query to $1-\delta$, which sets the coefficient to $\delta^{-2}$. Every arm was scored with
the Guo evaluator over 20 repetitions. The $C=10^6$ and $C=256$ arms are the uncapped and weight-capping
students of \cref{tab:controls}. For $C\le4096$, and across the grid family, FID stays between $1.0$ and $1.07$, and it degrades
progressively at larger coefficients.
Above $4096$ the five metrics move together at every step, and the median pre-clip norm rises from
$6$ to $253$ while the unit-norm clip binds on all 800 logged updates of every arm: the coefficient
changes the relative supervision weights, and the same unit-norm clipping rule is applied across
configurations.

\FloatBarrier %
\section{Deployment cost}
\begin{figure}[p]
  \centering
  \includegraphics[width=\linewidth]{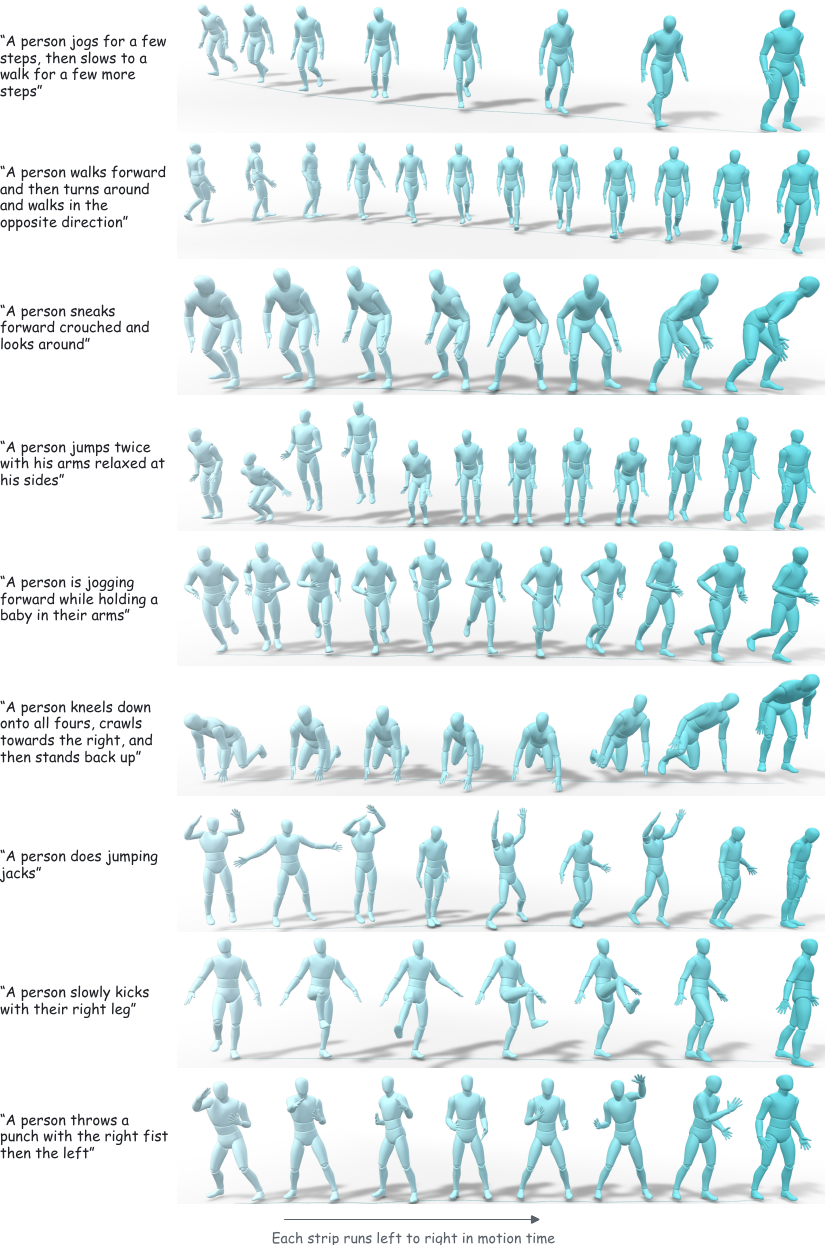}
  \caption{\textbf{Eight-step text-to-motion with TACD (I).} One prompt per row; each row is a
  single eight-step sample of \tacd\ (HY-Motion), the Qwen3-0.6B/Lite student, shown at evenly
  spaced time points (\cref{app:teaser}).}
  \label{fig:teaser}
\end{figure}
\begin{figure}[p]
  \centering
  \includegraphics[width=\linewidth]{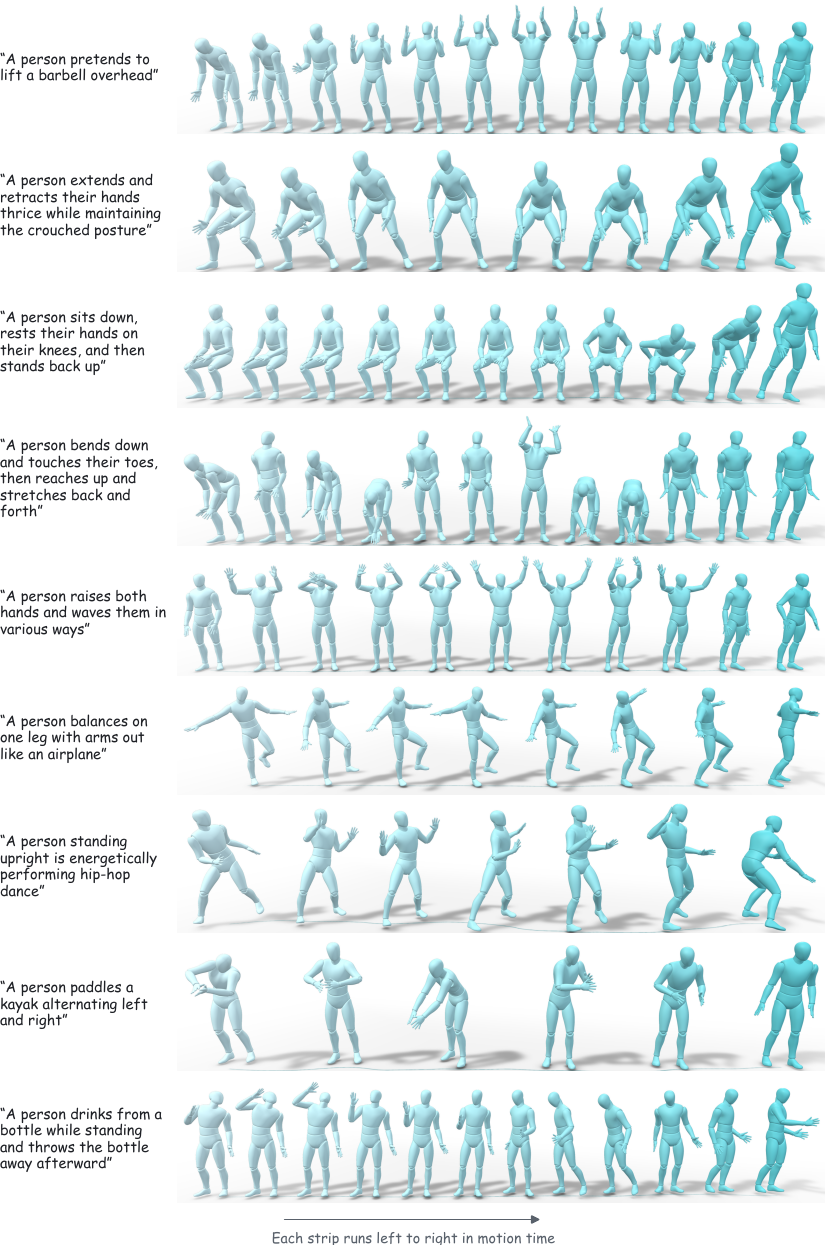}
  \caption{\textbf{Eight-step text-to-motion with TACD (II).} Continued from \cref{fig:teaser};
  same student, same rendering.}
  \label{fig:teaser_b}
\end{figure}

\label{sec:latency}
This appendix documents the latency and memory measurements behind \cref{fig:cost} and
\cref{sec:coupled} (\cref{app:latency}).
\subsection{Latency and memory}
\label{app:latency}
The batch-1 cost numbers in the paper (\cref{fig:cost,tab:kimodo_main,tab:efficiency,tab:kimodo,%
tab:latency}) come from one rig and one protocol: an RTX L40, \texttt{bf16} weights for the denoiser and
the text encoder, batch 1, all weights resident, no disk I/O in the timed loop, and
\texttt{cuda.synchronize} around text encoding and around sampling separately. HY-Motion rows generate a
196-frame ($6.5$\,s at 30\,fps) motion at CFG 5, Kimodo rows an eight-second motion. Each row runs 8
warm-up and 30 timed motions per pass, five passes in round-robin over all rows so that load drift on the
shared machine is spread over every row, and the reported latency is the median of the 150 timings.
Peak memory is the allocator's peak after loading, for text encoding plus one sample.

\cref{tab:latency} lists the HY-Motion rows. Text encoding costs 41--54\,ms per caption at batch 1, a small
portion of end-to-end latency. Sampling tracks the step
count: \tacd\ (HY-Motion) at eight steps samples in $227$\,ms vs.\ $1422$\,ms for 50-step Lite ($6.3\times$
for a $6.25\times$ step reduction). End to end, \tacd\ (HY-Motion) is $5.5\times$
faster than 50-step Lite and $7.7\times$ faster than the teacher, at $2.62$\,GB against the teacher's
$17.63$\,GB ($6.7\times$); the saving is the text encoder (8B to 0.6B) and the backbone (1B to 0.46B),
and the student adds only the $1024{\to}4096$ conditioning projection to the released Lite architecture.

\begin{table}[htbp]
\centering
\apptab
\begin{tabular}{@{}lrrrr@{}}
\toprule
Model / steps & end to end & sampling & text enc. & peak GB \\
\midrule
1B teacher $+$ Qwen3-8B, 50 steps                    & $2077$\,ms & $2026$ & $54$ & $17.63$ \\
0.46B Lite $+$ Qwen3-8B, 50 steps                    & $1476$\,ms & $1422$ & $52$ & $16.54$ \\
\rowcolor{methodwash}
\tacd\ (HY-Motion): 0.46B $+$ Qwen3-0.6B, 8 steps    & $268$\,ms  & $227$  & $41$ & $2.62$ \\
\bottomrule
\end{tabular}
\caption{\textbf{Batch-1 cost of the HY-Motion systems} on one RTX L40 (\texttt{bf16}; medians over five
round-robin passes of 30 motions of $6.5$\,s). Text encoding is per caption.}
\label{tab:latency}
\end{table}

\paragraph{Training cost.} Per iteration, on-policy supervision adds one $K$-step rollout of the
rollout-EMA copy, $K$ gradient-carrying student evaluations at the detached segment starts, and $KM$
frozen-teacher queries. All of this is training-only: inference NFE is unchanged, and the EMA copy does
not affect deployment.

\FloatBarrier %
\section{Qualitative results}
\label{app:qualitative}
This appendix shows generated motions: the teacher, its eight-step truncation and \tacd\ side by side
(\cref{app:teaser3}), and a gallery of eight-step samples of the compact student (\cref{app:teaser}).
\subsection{Teacher, truncated teacher, and TACD}
\label{app:teaser3}
\cref{fig:teaser3} shows four prompts, each rendered for the 50-step teacher, for the same teacher truncated
to the eight-step sampling grid, and for TACD at eight steps. Within a row the prompt, the
initial noise, the frame indices and the camera are shared: the three cells of a row come from a
single render whose camera is fitted to the union of the three motions.

\subsection{Eight-step gallery}
\label{app:teaser}
\cref{fig:teaser,fig:teaser_b} show one prompt per row. Each row is one eight-step sample of the
\tacd\ (HY-Motion), one draw per prompt. Within a row the poses are evenly spaced
time points of that one clip. For visualization only, each clip is rotated by a fixed yaw so that the body
faces the reader; this does not alter the motion.
\FloatBarrier

\end{document}